%% file: main.tex
\documentclass[11pt]{article}
\usepackage[a4paper,top=20mm,bottom=21mm,left=24mm,right=24mm]{geometry}
\usepackage[authoryear,round]{natbib}
\usepackage{fontspec,needspace,placeins}
\usepackage{amsmath,booktabs,tabularx,array,graphicx,float,caption,enumitem,xcolor,pdflscape,hyperref,fancyhdr}
\definecolor{teal}{HTML}{226786}
\hypersetup{colorlinks=true,linkcolor=teal,urlcolor=teal,citecolor=teal,pdftitle={GlobalHRBench: Does Changing the Question Language Change the Legal Answer?},pdfauthor={Yuding Wang and Yuhang Yao}}
\newcommand{\stat}[1]{\csname ghb#1\endcsname}
\input{tables/values}
\input{tables/archive_language_contrast}
\begin{document}\thispagestyle{plain}
\begin{center}{\LARGE\bfseries GlobalHRBench: Does Changing the Question Language\\[2pt] Change the Legal Answer?}\par
\vspace{4pt}{\large Yuding Wang\textsuperscript{1} \qquad Yuhang Yao\textsuperscript{2}}\\
{\small \textsuperscript{1}Farlinko \qquad \textsuperscript{2}Carnegie Mellon University}\end{center}
\begin{abstract}
\input{archive_abstract}
\end{abstract}
\section{Introduction}
\input{archive_introduction}

\section{Related Work}
\input{archive_related_work}

\section{Dataset Construction}\label{sec:data}
\subsection{Legal sources and question families}
\textbf{One decision per family.} Each family starts with a decision under one jurisdiction’s employment law. We documented the controlling original-language law, regulatory material or decision, its scope, and the date on which the rule was to be applied. Labour-law databases supported discovery and cross-checking. The legal reference date is 15 November 2026 for 209 families and 18 November for 31. Where an event date affects the result, the case supplies that date or the key specifies how to handle its absence.

\begin{figure}[!t]\centering
\makebox[\linewidth][c]{\includegraphics[width=178mm]{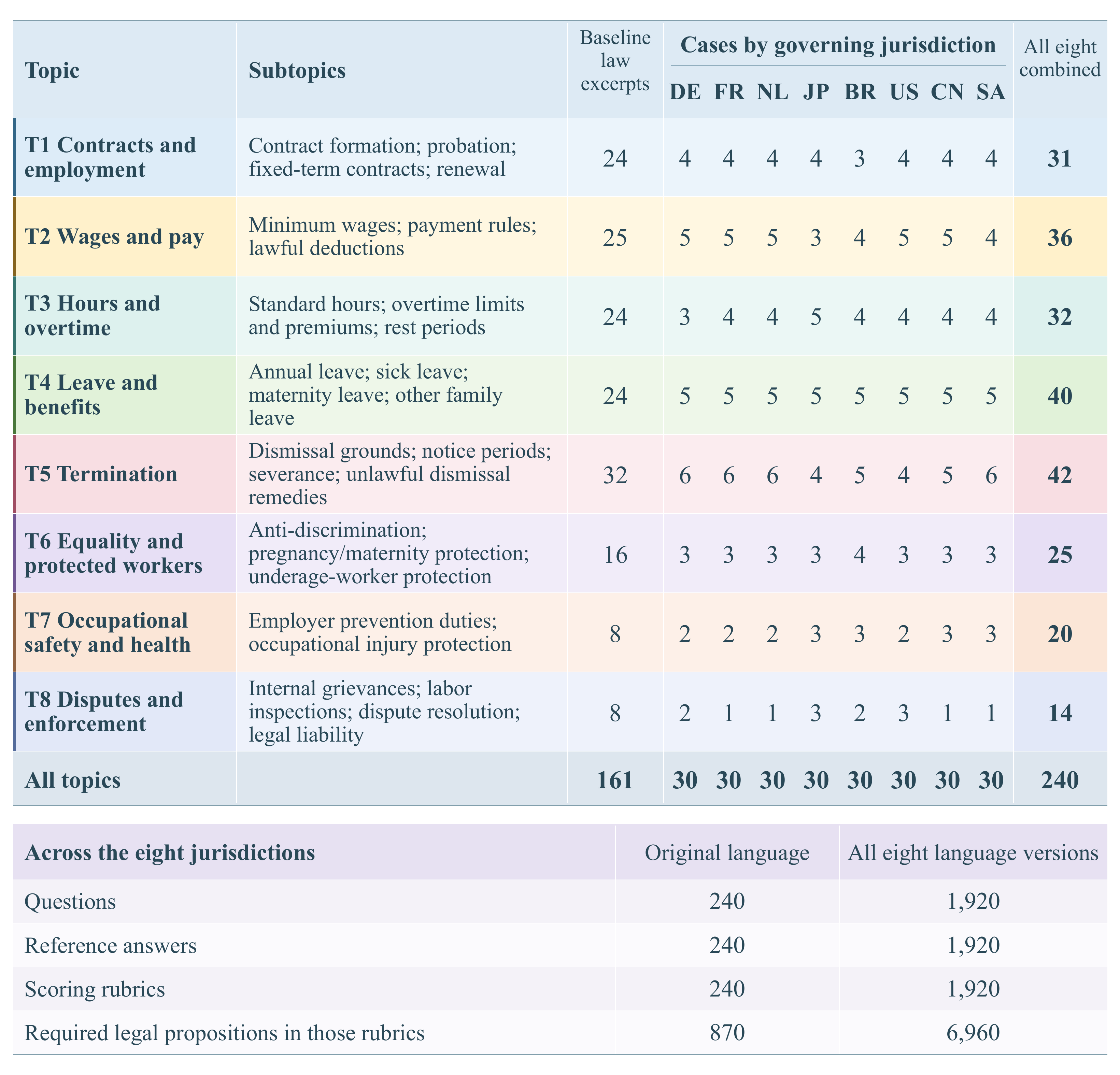}}
\caption{Contents of the frozen benchmark. The 161 excerpts count the baseline law collection. Some challenge questions also draw on additional authorities. A question can require several excerpts. Each family counts once under its primary HR topic and yields eight prompts, reference answers and scoring keys. A scoring key can require several legal propositions.}\label{fig:allocation}\end{figure}

\subsection{Sampling topics, difficulty and question format}
\textbf{Equal numbers by tier and format.} Each jurisdiction contributes 30 families. Within each of the Easy, Medium and Hard tiers, five are direct questions and five are scenarios. A direct question asks for a legal determination. A scenario supplies a sequence of facts or a proposed HR action and asks the respondent to decide what follows. Either format can be Easy. An Easy case can require applying a single rule or performing a short calculation. Medium cases require qualifications, calculations or interacting rules and passed native-language screening. A Hard case qualified after at least one verified substantive error during screening. Scores on the Hard tier therefore describe a deliberately selected challenge set.

\textbf{Shared topic coverage.} The same eight HR topics guided legal sampling and task selection in every jurisdiction. A \emph{provision} here means a legal rule or statutory excerpt selected as source material. A \emph{question family} is one legal problem and its eight language versions; it counts once. A case may depend on several rules. The blueprint gives more slots to topics with more distinct decisions, such as termination, than to narrower topics such as enforcement. We fixed 30 families per jurisdiction and five per difficulty--format combination. After selecting the Hard cases, we filled topic gaps with eligible Medium and Easy cases. Figure~\ref{fig:allocation} counts the legal excerpts and the question families that entered the benchmark.

\textbf{Coverage in the frozen set.} Every jurisdiction covers all eight topics, including five leave-and-benefit cases and four to six termination cases. This sampling gives every model the same spread of legal decisions in every language. The topic mix represents the HR decisions covered by the benchmark. Estimating their frequency in practice would require a sample of workplace requests. Appendix Table~\ref{tab:allocationlaw} illustrates the decisions covered by each topic and gives their numbers at each difficulty.

\subsection{Translation and scoring keys}
\textbf{Eight versions of each case.} The source languages are German (Germany), French (France), Dutch (the Netherlands), Japanese (Japan), Portuguese (Brazil), English (the United States), Chinese (mainland China) and Arabic (Saudi Arabia). Each original question and any scenario materials have seven translations. The gold answer and scoring key have corresponding language versions. Each of the eight prompt languages thus covers exactly the same 240 families.

\textbf{Legally equivalent translations.} Reviewers checked which side of a threshold qualifies, the dates that start and end a period, the calculation base and any exception. If a case deliberately omits a decisive fact, every version must preserve that omission, and the key must accept the legally valid branches. Only required propositions count towards completeness. A shorter, legally equivalent answer can receive full credit. Appendix~\ref{app:audit} documents the review and the revisions to the frozen questions.

\section{Main Experiment: Closed-Book Evaluation}\label{sec:mainsetup}
\subsection{Models and first-answer protocol}
\textbf{One independent answer per prompt.} We evaluate GPT-6.1 Sol, GPT-6 Luna (requested snapshot 2026-09-22, returned alias GPT-6 Luna), Claude Sonnet 5.5, Kimi K3.1 Pro, Gemini 3.5 Flash-Lite, GLM-5.3, DeepSeek V4.1 Flash and two Gemma 4 models (31B and 26B A4B). Each model receives all 1,920 prompts. It sees only the question, any scenario materials and the response instructions, and must answer in the requested language in one independent attempt. The protocol excludes supplied law, search, tools and feedback. All 17,280 planned positions enter the end-to-end denominator. A transport failure triggers a check for an existing response to the identical request and, where needed, a technical resend. We score the first observed model response, including any observed search. A truncated response is scored on the text it contains. A cap-exhausted attempt with no answer cannot pass.

\begin{figure}[!htbp]\centering
\makebox[\linewidth][c]{\includegraphics[width=178mm]{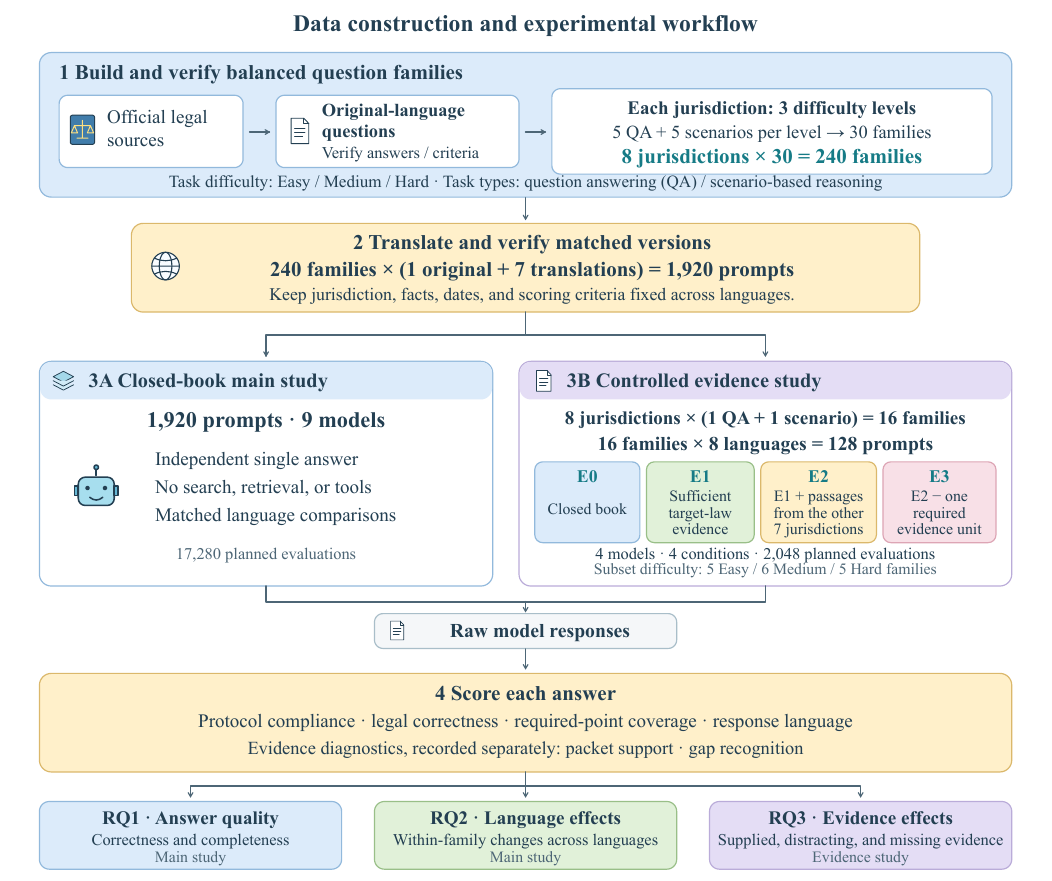}}
\caption{The diagnostic sample reuses all 512 closed-book (E0) responses from the corresponding main-study positions. Within each family the supplied legal excerpts and their order are identical across prompt languages in a given condition.}\label{fig:design}\end{figure}

\Needspace{7\baselineskip}
\textbf{Two uses of the same cases.} Figure~\ref{fig:design} shows how the experiments share the closed-book baseline. The main experiment compares language versions within a family. The ablation in Section~\ref{sec:evidence} returns to 16 of those families, retains their closed-book responses as E0 and changes the evidence supplied with the same prompt.

\subsection{Answer scoring}
\textbf{What earns a strict pass.} We score an eligible, substantive answer on three dimensions. \emph{Correctness} requires that its legal assertions contain no material error. \emph{Completeness} requires all propositions in the legal key, including decisive qualifications. \emph{Language compliance} requires use of the requested answer language. Strict pass requires all three. A short answer can be correct and complete. A longer one can supply a correct figure while misstating when the rule applies. Among the 385 retained cross-checked responses with matching identities and scoring keys, strict labels agreed before adjudication in 364 cases ($\kappa=0.886$). Appendix~\ref{app:audit} gives the sampling and grading history.

\subsection{Paired comparisons and uncertainty}
\textbf{Pairing within a family.} We compare one model’s answers to the same legal question in two languages. For a French case, French--German is one pair and French--Arabic is another. The 240 families each yield seven original-to-translation comparisons, giving 1,680 pairs per model. An original pass followed by a translated failure is a \emph{loss}. An original failure followed by a translated pass is a \emph{gain}. The net change subtracts losses from gains, while the flip rate adds them. Both divide by all 1,680 pairs, including those that pass twice or fail twice. A budget-exhausted empty answer and an actual tool violation both fail the affected position.

\textbf{Family-level uncertainty.} Each original answer contributes to seven paired comparisons. We sample whole families with replacement, keeping their language versions and, in the diagnostic study, their evidence conditions together. Intervals are the 2.5th and 97.5th percentiles of 5,000 draws (seed 20261122). They describe variability across the sampled cases without adjustment for multiple comparisons.

\section{Main Results}\label{sec:mainresults}
\subsection{Which models deliver correct and complete advice?}
\textbf{Different failures lower the same score.} A model can give incorrect advice or exhaust its output allowance before delivering an answer (Table~\ref{tab:fail}). Sol leads at \stat{solstrict}/1,920 (\stat{solpct}\%) and usually delivers text. GLM produces text on most prompts, but \stat{glmlaw}/\stat{glmgrade} substantive, eligible answers are legally incorrect. DeepSeek V4.1 Flash exhausts its 2,080-token completion allowance on \stat{dsempty} prompts before any answer text appears. All 1,558 requests completed with a length stop; the provider attributed all 2,080 tokens to reasoning for 1,557 of them and 2,079 for the other. Among the \stat{dsgrade} substantive answers it does deliver, \stat{dscorrect} are legally correct. Its gradable subset is mostly Easy (\stat{dseasy}/\stat{dsgrade}). This selection matters for interpreting its conditional accuracy.

\begin{table}[!htbp]\centering\small\input{tables/archive_causes}
\caption{Strict pass rate and pass count appear first. The failure columns assign each remaining prompt to one cause, in priority order. Every technical incident was resolved against an identical-request model response. An observed tool call violates the closed-book protocol. A cap-empty response used its output allowance before producing an answer. A fragment contains too little answer text to grade. Among eligible texts, a legal error is a materially wrong claim, an omission is an otherwise correct answer missing a required point, and a language-only failure meets the full legal key in the wrong language. A transport error is checked against the original receipt and any identical-request recovery before assigning an outcome.}\label{tab:fail}\end{table}

\input{deepseek_budget_note}

\textbf{Legal accuracy on shared deliveries.} On the \stat{matchn} prompts on which both models delivered substantive answers, DeepSeek is legally correct \stat{matchd} times and GLM \stat{matchg} times. On these shared prompts \stat{matchdonly} favor DeepSeek alone and \stat{matchgonly} favor GLM alone. This comparison addresses content on matched deliveries. The end-to-end score also counts the other prompts, including those on which the model produced no answer. Luna and Sonnet lose \stat{lunaomit} and \stat{snomit} positions, respectively, to missing required points after legal correctness is satisfied and before language is checked. Gemma 26B A4B records \stat{gsmallcap} cap-empty positions, compared with \stat{gbigcap} for Gemma 31B. This same-provider comparison also changes parameterization and training, so it cannot isolate an architecture effect.

\textbf{Rule application.} Reviewers frequently attributed GLM's legal errors to applying the wrong rule; Sonnet also misapplied rules to the facts or chose the wrong operative date (Appendix Table~\ref{tab:tags}). Reasoning length and elapsed time measure different costs. Median main-request times were 11.9 seconds for DeepSeek, 21.8 for Sol, 49.9 for Kimi and 3.8 for GLM.

\subsection{Which cases change outcome across languages?}
\begin{figure}[!htbp]\centering\includegraphics[width=.98\linewidth]{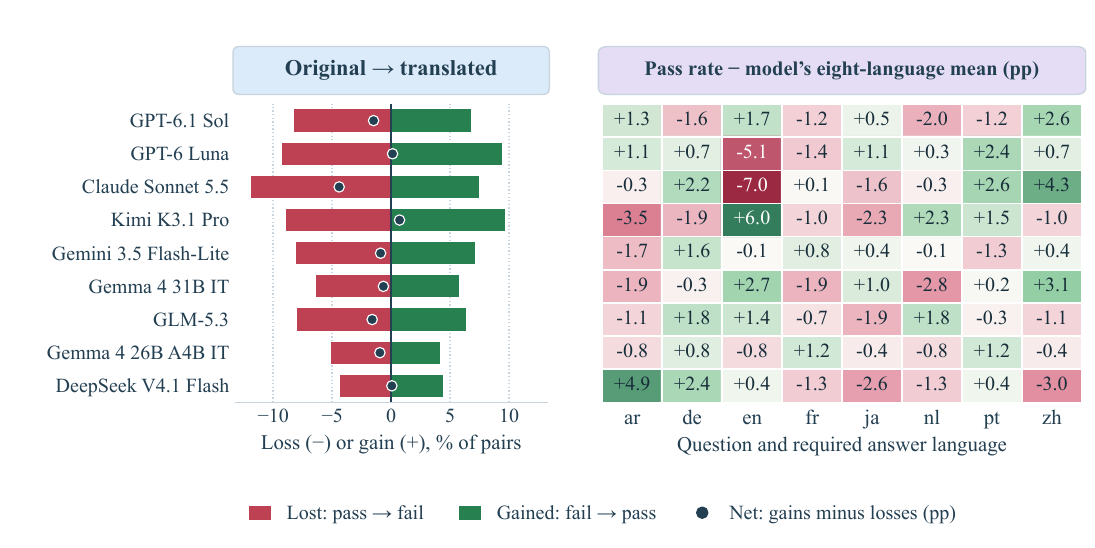}
\caption{Read each row of the left panel from zero. The pink bar extending left counts original passes lost on translation as a share of all 1,680 pairs. The green bar extending right counts gains on the same scale. The dot marks the net difference in percentage points. For Kimi, 150 losses and 162 gains place the dot just 12 pairs above zero. Right: a signed cell gives the language's strict pass rate minus that model's own eight-language average, in percentage points. A positive cell means above that model's average. Each language uses all 240 cases and is also the required answer language.}\label{fig:switches}\end{figure}

\textbf{Gains and losses can cancel.} Kimi loses \stat{kimilost} passes on translation and gains \stat{kimigain}. These two counts yield a net difference of \stat{kimipp} percentage points and a flip rate of \stat{kimiflip}\%. Sonnet's losses exceed its gains by \stat{sonnetnet} pairs. The direction of the change depends on the model (Figure~\ref{fig:switches}). In English, Kimi passes \stat{kimien}/240 (46.7\%), 6.0 percentage points above its own eight-language average; Sonnet passes \stat{sonneten}/240 (37.5\%), 7.0 points below its own average. Chinese is among Sonnet's strongest versions. A language that performs well for one model can perform poorly for another.

\textbf{Translations can outperform the original.} Across 504 comparisons (nine models, eight jurisdictions and seven translations), the translated language gains passes on balance in 188, ties in 84 and loses passes in 232. Each comparison holds 30 families fixed and several comparisons share the same original answers. On the same 30 French-law cases, Gemini 3.5 Flash-Lite passes 10 in Chinese and seven in French. GPT-6 Luna also gains on balance (16 versus 14), while Claude Sonnet 5.5 passes fewer in Chinese (16 versus 17). These comparisons change the language of both the question and the requested answer. They describe observed outcomes for these families and models; identifying a preferred language for future cases would require a separate sample.

The response language explains much of the English--Chinese reversal between Kimi and Sonnet. In English, Kimi leads on strict passes (\KimiEnStrict{} versus \SonnetEnStrict{}); in Chinese, Sonnet leads (\SonnetZhStrict{} versus \KimiZhStrict{}). Holding delivery, protocol and the full legal key fixed but dropping only the answer-language requirement, Sonnet leads in both languages: \SonnetEnContent{} to \KimiEnContent{} in English and \SonnetZhContent{} to \KimiZhContent{} in Chinese. The swing in the two models' relative rates is \PairStrictDiff{} percentage points under strict scoring (95\% family-bootstrap interval \PairStrictLo{} to \PairStrictHi{}), and \PairContentDiff{} points under content scoring (\PairContentLo{} to \PairContentHi{}). We selected this contrast after examining the language results. Both scores use all 240 questions per language.

\begin{table}[!htbp]\centering\small\input{tables/pair}
\caption{Each model has 1,680 original-to-translation pairs. For Kimi, 312 pairs change outcome (150 losses plus 162 gains), or 18.6\%; the overall net gain is 12/1,680, or 0.71 percentage points. Only the last column conditions on an original pass, so its denominator varies with the model. Intervals resample families.}\label{tab:switch}\end{table}
\textbf{Stable failure also lowers the flip rate.} DeepSeek seldom passes in either language. Conditional on an original pass, it loses about three in ten comparisons. In \stat{dscapflip} switched pairs its failing side is empty at the cap, so no legal conclusion can be compared on that side. In contrast, Sonnet's losses include \stat{snlegalloss} incorrect legal answers, \stat{snomitloss} correct but incomplete answers and \stat{snlangloss} legally correct and complete answers in the wrong language. A single original answer can contribute to several pairs. Table~\ref{tab:switch} reports each model separately.

\textbf{The model ranking can reverse.} Table~\ref{tab:languagecounts} compares all nine models on the same cases. Kimi leads Sonnet by 22 passes in English, and Sonnet leads Kimi by 22 in Chinese. A second reversal appears for Gemma 4 31B and GLM: Gemma leads 58 to 39 in Chinese, while GLM leads 46 to 44 in Dutch. The two-case Dutch advantage is small. Arabic illustrates a different comparison. It produces DeepSeek's highest count and Kimi's lowest, yet Kimi still passes 89 Arabic prompts to DeepSeek's 45. Sol has the highest absolute count in every language on this sample.

\begin{table}[!htbp]\centering\small\setlength{\tabcolsep}{3.3pt}\renewcommand{\arraystretch}{1.12}
\input{tables/archive_language_counts}
\caption{Strict passes on the same 240 cases in each language. Bold marks the highest observed count within that model, and underlining marks its lowest, including ties. Read down a column to compare models for a working language. The last column gives the decline from the best language to the worst relative to the best, $100(H-L)/H$, where $H$ and $L$ are the highest and lowest pass counts in that row. Port. denotes Portuguese.}\label{tab:languagecounts}\end{table}

\textbf{Two routes to a lost pass.} A change in answer language or a change in legal reasoning can cost a pass. On the English version of the French rest-break item (GHB-F-FR-09), Sonnet wrote in French that reaching six hours triggers at least twenty consecutive minutes. It satisfied both legal propositions in both languages. The English version failed solely on the answer-language requirement. On another French case (GHB-EX-FR-Q46), a worker accepted a professional security contract on 13 July, employment ended on 24 July, and the worker both contested the economic grounds and sought priority for rehire on 19 July the following year. The French answer kept the deadlines separate. The English answer counted from termination for the challenge and concluded “the deadline does not bar it.” Both versions used their requested language.\footnote{The two clocks run from acceptance and termination, respectively. See French Labour Code, \href{https://code.travail.gouv.fr/code-du-travail/l1233-67}{Art. L.1233-67} and \href{https://code.travail.gouv.fr/code-du-travail/l1233-45}{Art. L.1233-45}.} The latter pair records a change in rule application. Each version was sampled once, so neither pair establishes the cause of the discrepancy.

\subsection{How does performance vary by jurisdiction?}
\textbf{Jurisdiction effects vary by model.} Table~\ref{tab:country} compares jurisdictions within each model. Kimi passes \stat{kimicn}/240 on Chinese law and \stat{kimide}/240 on German law. Sonnet passes \stat{sonnetjp}/240 on Japanese law and \stat{sonnetsa}/240 on Saudi law. Each column describes a different set of 30 problems. Differences can arise from the selected facts, areas of law or model knowledge. Appendix~\ref{app:regional} instead fixes the 30 problems within each jurisdiction and shows how many more or fewer cases pass in each language for each model.
\begin{table}[!htbp]\centering\small\input{tables/country_heatmap}
\caption{Strict pass rate (\%) by governing jurisdiction. Each cell is the absolute percentage passing among 30 cases answered in eight languages (240 answers per model). Darker cells indicate higher pass rates on a common scale. Compare across a row for one model or down a column for one jurisdiction.}\label{tab:country}\end{table}

\subsection{How do language outcomes vary with difficulty and question format?}
\textbf{Agreement depends on difficulty.} Sol changes pass status on \stat{solflipe}\% of its Easy pairs and \stat{solfliph}\% of its Hard pairs. DeepSeek shows the opposite pattern (\stat{dsflipe}\% and \stat{dsfliph}\%). A pair in which reasoning exhausts the budget on both versions agrees on failure. Across all eight language versions, DeepSeek has \stat{dsnone} families without a pass and \stat{dsall} with eight passes. Sonnet has \stat{snpurefail} and \stat{snpurepass}, respectively. Consistent failure can resemble language robustness if only the flip rate is reported.

\begin{figure}[!htbp]\centering\includegraphics[width=.98\linewidth]{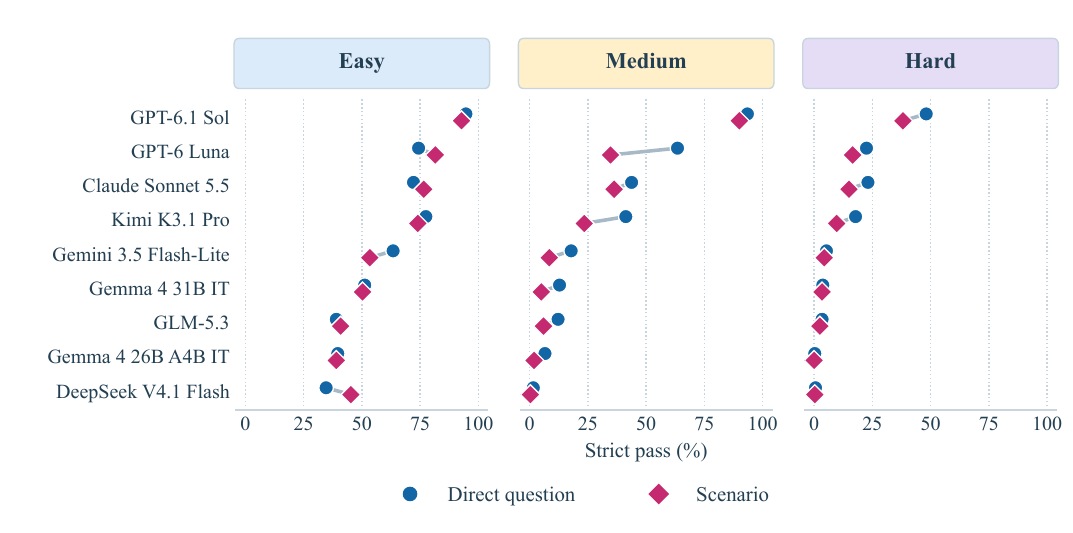}
\caption{Blue circles show direct questions; pink diamonds show scenarios. Each point is a strict pass rate on 40 families in eight languages (320 positions). The connector compares the two formats at the same difficulty for one model; the underlying cases differ by format.}\label{fig:tier}\end{figure}

\textbf{Scenario effects vary by tier.}
Luna passes more Easy scenarios than Easy direct questions (261 versus 238 of 320), but the ordering reverses at Medium difficulty (111 versus 203 of 320; Figure~\ref{fig:tier}). The Medium gap is \stat{lunafmtgap} percentage points. Kimi's Medium direct questions do best in English and Portuguese. Its Medium scenarios reach \stat{kimimedscmax}/40 in English and \stat{kimimedscmin}/40 in Arabic. Sonnet's Easy scenarios range from \stat{sonneteasyen}/40 in English to \stat{sonneteasyde}/40 in German. Each 40-family cell uses the same cases across languages. Extrema selected from many such cells warrant replication. Full counts, ties and component scores for every language, tier and form accompany the paper.

\textbf{Longer scoring keys.} Easy QA keys contain two or three required propositions. Hard scenario keys range from three to fourteen, with five cases requiring at least six. On Hard items Sol earns \stat{solpointsh}\% of required points across gradable texts, while only \stat{solhardpass}/640 pass strictly. On Medium items it earns \stat{solpointsm}\% of points. GLM's corresponding Hard coverage is \stat{glmpointsh}\%. This micro-average counts rubric points on gradable responses. It excludes empty attempts and does not isolate the effect of longer rubrics.

\textbf{From closed book to supplied evidence.} The main experiment observes each model's answer without access to the governing text. Its failure counts distinguish wrong advice, omitted qualifications and reasoning that uses the completion allowance before yielding advice. They leave open which of these failures would change if the relevant law were supplied. We examine that question on a paired subset below.

\input{evidence}
\FloatBarrier
\section{Takeaways and Recipes}
\input{archive_takeaways}
\section{Conclusion}
\input{archive_conclusion}
\section*{Limitations}
\input{archive_limitations}
\bibliographystyle{acl_natbib}
\bibliography{references}

\appendix
\input{audit_appendix}
\begingroup\renewcommand\small{\fontsize{9.8}{11.4}\selectfont}\setlength{\parskip}{2pt plus .5pt}\setlist[enumerate]{itemsep=1pt,topsep=2pt,parsep=0pt}
\makeatletter\renewcommand\paragraph{\@startsection{paragraph}{4}{0pt}{7pt plus 1pt}{-0.7em}{\normalfont\normalsize\bfseries}}\makeatother
\input{example}\endgroup\normalsize
\newgeometry{top=13mm,bottom=13mm,left=12mm,right=12mm,headheight=0pt,headsep=0pt,footskip=0pt}
\begin{landscape}
\thispagestyle{empty}
\noindent{\small\sffamily GlobalHRBench\hfill Paired multilingual evaluation of employment-law QA}\par
\vspace{3pt}\hrule height .3pt\vspace{1pt}
\section{Within-jurisdiction language comparisons}\label{app:regional}
Figure~\ref{fig:reg} compares each translated version with the original language on the same 30 cases. Working in the language of the governing law does not uniformly give the best results. On the same 30 Chinese-law cases, Kimi has 21 passes in English and 18 in Chinese, whereas Sonnet has 13 in English and 15 in Chinese. Each cell shows a net change against the original language; one case corresponds to 3.33 percentage points. A zero can also arise when individual gains and losses cancel.
\begin{figure}[!htbp]\centering
\includegraphics[width=.99\linewidth,height=125mm,keepaspectratio]{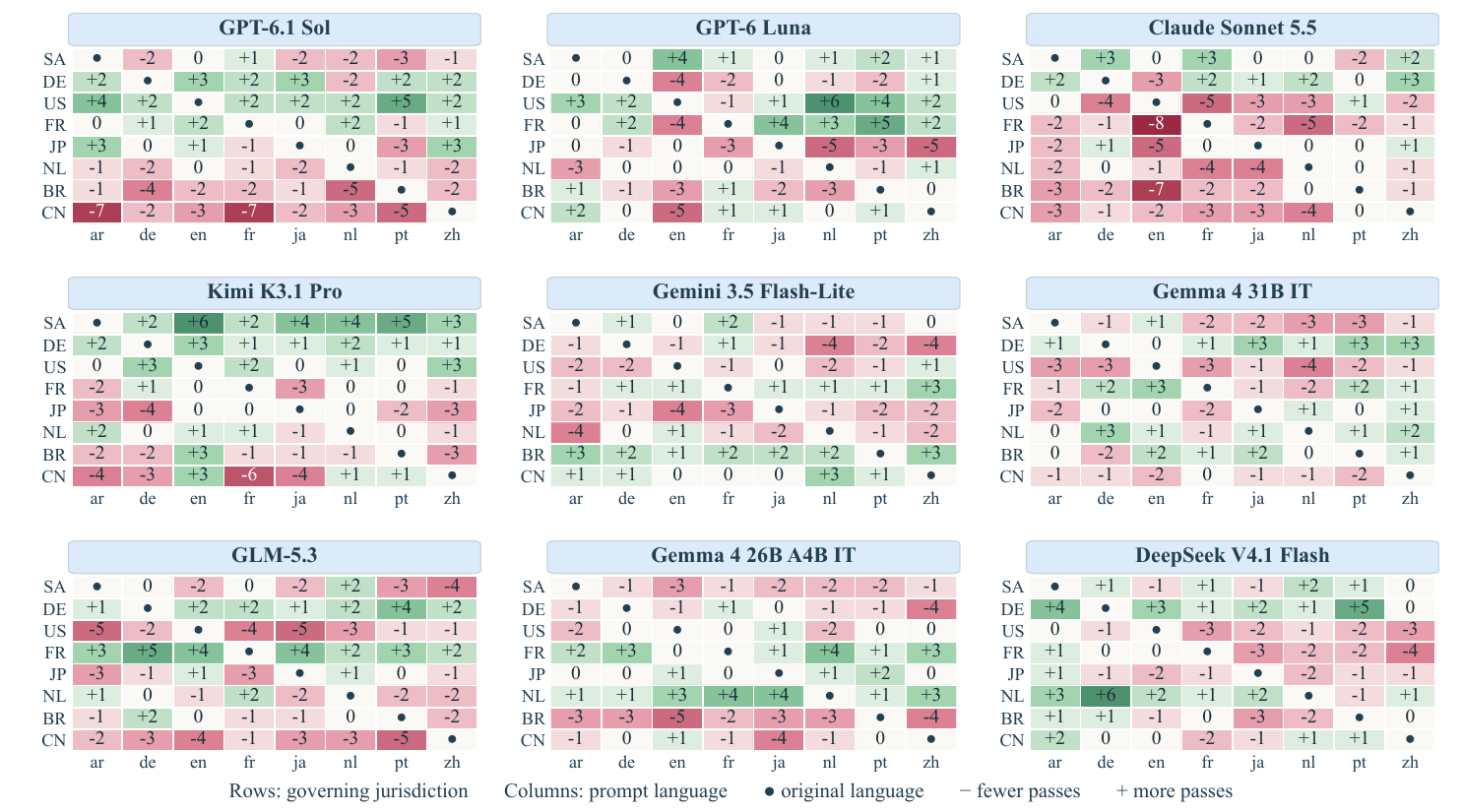}
\caption{Change in the number of strict passes. Each panel is one model, and each row is one governing jurisdiction. The original-language cell is marked with a dot.}\label{fig:reg}\end{figure}
\vfill\centerline{\small\thepage}\end{landscape}
\end{document}

%% file: tables/values.tex
\expandafter\def\csname ghbdsall\endcsname{9}
\expandafter\def\csname ghbdscapflip\endcsname{116}
\expandafter\def\csname ghbdscorrect\endcsname{301}
\expandafter\def\csname ghbdseasy\endcsname{329}
\expandafter\def\csname ghbdsempty\endcsname{1558}
\expandafter\def\csname ghbdsflipe\endcsname{23.9}
\expandafter\def\csname ghbdsfliph\endcsname{1.2}
\expandafter\def\csname ghbdsgrade\endcsname{362}
\expandafter\def\csname ghbdsnone\endcsname{178}
\expandafter\def\csname ghbgbigcap\endcsname{358}
\expandafter\def\csname ghbglmgrade\endcsname{1853}
\expandafter\def\csname ghbglmlaw\endcsname{1370}
\expandafter\def\csname ghbglmpointsh\endcsname{35.7}
\expandafter\def\csname ghbgsmallcap\endcsname{1005}
\expandafter\def\csname ghbkimicn\endcsname{132}
\expandafter\def\csname ghbkimide\endcsname{75}
\expandafter\def\csname ghbkimien\endcsname{112}
\expandafter\def\csname ghbkimiflip\endcsname{18.6}
\expandafter\def\csname ghbkimigain\endcsname{162}
\expandafter\def\csname ghbkimilost\endcsname{150}
\expandafter\def\csname ghbkimimedscmax\endcsname{13}
\expandafter\def\csname ghbkimimedscmin\endcsname{6}
\expandafter\def\csname ghbkimipp\endcsname{+0.71}
\expandafter\def\csname ghblunafmtgap\endcsname{28.8}
\expandafter\def\csname ghblunaomit\endcsname{305}
\expandafter\def\csname ghbmatchd\endcsname{300}
\expandafter\def\csname ghbmatchdonly\endcsname{93}
\expandafter\def\csname ghbmatchg\endcsname{224}
\expandafter\def\csname ghbmatchgonly\endcsname{17}
\expandafter\def\csname ghbmatchn\endcsname{361}
\expandafter\def\csname ghbsnlangloss\endcsname{26}
\expandafter\def\csname ghbsnlegalloss\endcsname{138}
\expandafter\def\csname ghbsnomit\endcsname{131}
\expandafter\def\csname ghbsnomitloss\endcsname{35}
\expandafter\def\csname ghbsnpurefail\endcsname{78}
\expandafter\def\csname ghbsnpurepass\endcsname{34}
\expandafter\def\csname ghbsolflipe\endcsname{9.6}
\expandafter\def\csname ghbsolfliph\endcsname{23.6}
\expandafter\def\csname ghbsolgraded\endcsname{1876}
\expandafter\def\csname ghbsolhardpass\endcsname{276}
\expandafter\def\csname ghbsolpct\endcsname{76.20}
\expandafter\def\csname ghbsolpointsh\endcsname{78.0}
\expandafter\def\csname ghbsolpointsm\endcsname{98.9}
\expandafter\def\csname ghbsolstrict\endcsname{1463}
\expandafter\def\csname ghbsonneteasyde\endcsname{38}
\expandafter\def\csname ghbsonneteasyen\endcsname{20}
\expandafter\def\csname ghbsonneten\endcsname{90}
\expandafter\def\csname ghbsonnetjp\endcsname{123}
\expandafter\def\csname ghbsonnetnet\endcsname{74}
\expandafter\def\csname ghbsonnetsa\endcsname{78}

%% file: tables/archive_language_contrast.tex
\newcommand{\KimiEnStrict}{112}
\newcommand{\KimiZhStrict}{95}
\newcommand{\SonnetEnStrict}{90}
\newcommand{\SonnetZhStrict}{117}
\newcommand{\PairStrictDiff}{18.3}
\newcommand{\PairStrictLo}{9.2}
\newcommand{\PairStrictHi}{27.9}
\newcommand{\KimiEnContent}{113}
\newcommand{\KimiZhContent}{101}
\newcommand{\SonnetEnContent}{118}
\newcommand{\SonnetZhContent}{117}
\newcommand{\PairContentDiff}{4.6}
\newcommand{\PairContentLo}{-2.9}
\newcommand{\PairContentHi}{12.5}

%% file: archive_abstract.tex
Language matters when deploying LLM agents across borders, especially in law, where the governing rules must hold across working languages. Does changing the language of a legal question change answer accuracy? We introduce GlobalHRBench, pairing 240 employment-law problems across eight jurisdictions and eight languages while fixing facts, applicable law and required conclusions. We evaluate nine models on all 1,920 prompts in independent, closed-book, single-answer trials, scoring legal correctness, completeness and answer-language compliance. GPT-6.1 Sol leads at 76.2\% strict passes, with a 5.8\% relative best-to-worst language decline, compared with 42.2\% for DeepSeek V4.1 Flash. Kimi K3.1 Pro and Claude Sonnet 5.5 reverse order between English and Chinese. A four-model study on 16 paired families compares closed-book answers with governing evidence, added cross-jurisdiction distractors and removal of a governing passage. Governing evidence corrects legal errors and shortens reasoning on recovered answers, yet recognizing an evidence gap does not ensure correct application of the rules that remain. These findings motivate multilingual training for accurate, efficient and evidence-supported answers, and deployment evaluation in the user's working language.

%% file: archive_introduction.tex
\begin{figure}[!t]\centering\makebox[\linewidth][c]{\includegraphics[width=166mm]{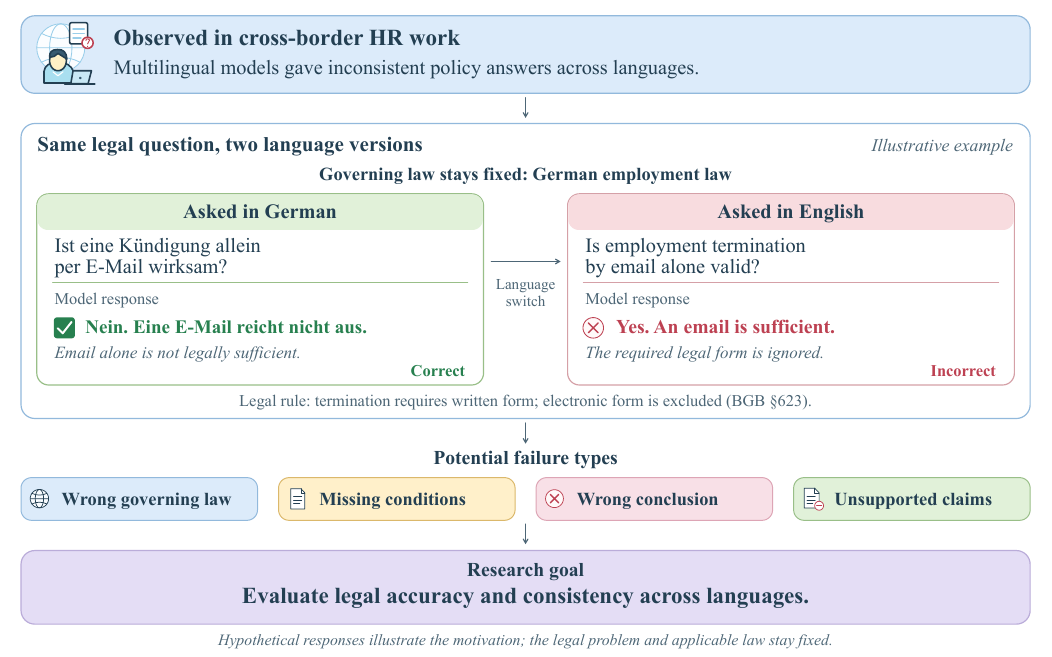}}
\caption{The two answers are hypothetical. The governing rule is \href{https://www.gesetze-im-internet.de/bgb/__623.html}{§623 of the German Civil Code}. The benchmark tests whether differences of this kind occur in recorded model answers.}\label{fig:motivation}\end{figure}

Employment-law questions routinely cross language boundaries. A worker, an HR manager and a legal adviser may discuss the same employment decision in different languages while remaining subject to the same rules. A model serving these users must preserve the applicable conditions, exceptions and remedies as it changes the language of its answer. For example, a question about terminating a German employment contract by email still requires a determination under German law when asked in English. Fluent wording alone gives the reader little assurance that the model has preserved that determination. Figure~\ref{fig:motivation} illustrates how legal errors, omissions and unsupported conclusions can affect the same decision.

Does asking the same legal question in another language change the accuracy of the answer? Comparing French questions about French law with Japanese questions about Japanese law changes the governing rules and the questions at the same time. Differences in topic, factual complexity or required qualifications can then appear as differences in language performance. Multilingual legal benchmarks broaden linguistic coverage, and paired evaluations have examined legal multiple-choice items and bilingual labour-law QA \citep{niklaus,alsagheer,chi}. Employment advice adds the challenge of open-ended generation. A response must state the conclusion and the qualifications necessary to act on it, including conditions the question does not explicitly ask the model to enumerate.

We introduce GlobalHRBench, comprising 240 employment-law question families from eight jurisdictions. Each family has eight language versions with aligned facts, applicable dates, governing law and required conclusions, yielding 1,920 prompts. Every jurisdiction contributes 30 families, with five direct questions and five scenarios at each of three difficulty levels. A shared framework covers eight employment-law topics, with topic gaps filled during selection. The resulting design allows comparison across languages on exactly the same legal decisions and examination of how those differences vary by jurisdiction, difficulty and question format.

We evaluate nine models in independent, closed-book, single-answer trials. A strict pass requires legally correct and complete advice in the requested language, with a valid response under the protocol. All 17,280 planned positions enter the denominator, including attempts that exhaust their output allowance before delivering an answer. We also score correctness, coverage of required propositions and language compliance separately. These measurements let us distinguish a mistaken legal determination, an omitted qualification, a sound answer in the wrong language and a failure to deliver text. The distinctions matter when deciding what intervention might improve the result.

GPT-6.1 Sol has the highest overall strict pass rate, 76.2\%, and the smallest relative best-to-worst language decline, 5.8\%. The corresponding decline reaches 42.2\% for DeepSeek, from 45 to 26 passes on the same 240 problems. These are differences in end-to-end success under the tested output allowance. A translated question can outperform its original-language version, and the preferred model can change with the working language. Kimi leads Sonnet in English by 22 passes, while Sonnet leads by 22 in Chinese. Sonnet leads on legally correct, complete content in both languages. Its strict English score loses 28 otherwise passing answers because they use another language. The distinction exposes a deployment-relevant failure that a content-only ranking would miss.

On 16 shared families, we evaluate four models under closed-book, governing-evidence, added-distractor and evidence-removal conditions. This diagnostic design preserves the legal answer key while changing what the model can establish from the supplied material. It reveals improvements with different sources. Governing passages and evidence-use instructions correct material legal errors in 49 of Sonnet's 70 newly passing responses. For DeepSeek, all 21 prompts that recover from cap-exhausted empty answers to passes use fewer reasoning tokens. Removing a passage reveals another weakness. Sonnet can identify the missing sick-pay rule yet misapply the separate hours limit still present in the packet.

\textbf{Contributions.} We make three contributions to multilingual legal evaluation.
\begin{itemize}[leftmargin=1.4em,topsep=2pt,itemsep=4pt,parsep=0pt]
\item \textbf{An aligned employment-law resource.} GlobalHRBench provides \textbf{240 families across eight jurisdictions and eight languages}, comprising 1,920 prompts with reference answers and scoring keys. The original keys contain 870 required legal propositions, aligned across the language versions, with equal family counts by jurisdiction, difficulty and format.
\item \textbf{A model-specific account of language sensitivity.} Across \textbf{nine models and 17,280 evaluation positions}, we measure relative language gaps up to 42.2\%, ranking reversals and the distinct contributions of legal errors, omissions and language compliance. Paired comparisons retain the same families, and uncertainty estimates resample whole families.
\item \textbf{A diagnostic of what evidence changes.} Across \textbf{2,048 positions covering four models, 16 families and four evidence conditions}, we identify legal corrections, reasoning savings and failures on still-supported conclusions. Fixed legal keys and separate source-support scores distinguish these effects.
\end{itemize}

%% file: archive_related_work.tex
\textbf{Legal tasks and jurisdictions.}  LexGLUE evaluates English legal-language tasks \citep{chalkidis}, LawBench tests Chinese legal knowledge at three cognitive levels \citep{fei}, and LegalBench distinguishes six forms of legal reasoning \citep{guha}. LEXTREME broadens coverage to 11 datasets in 24 languages \citep{niklaus}. For employment law, LaborBench uses U.S. Department of Labor comparisons to test whether systems can recover differences in unemployment insurance across states \citep{hariri}. PLawBench scores practical legal consultations and case analyses against expert-designed rubrics \citep{plaw}. Our unit of comparison is a single employment decision, with the same legal requirements for every language in which a user asks about it.

\textbf{Language-dependent rankings.} \citet{kreutzer2025dejavu} found that model rankings on a European multilingual leaderboard change with the languages and tasks emphasized. GlobalHRBench tests how the ordering changes within employment law when the cases and scoring requirements are fixed across languages.

\textbf{Paired multilingual questions.} How questions are collected determines what a language comparison can establish. TyDi QA elicits questions independently in each language \citep{tydi}. MLQA aligns extractive questions across languages \citep{mlqa}, MKQA aligns open-domain questions to language-independent answers \citep{mkqa}, and Belebele holds multiple-choice reading items fixed \citep{belebele}. XOR QA asks systems to find evidence across languages \citep{xor}. In law, bilingual BSARD pairs French and Dutch statutory retrieval, including translated questions \citep{bsard}. \citet{alsagheer} hold 200 legal multiple-choice items fixed across English and three Arabic varieties. \citet{chi} evaluate 231 Vietnamese--English labour-law pairs with retrieval and verifier-guided correction. We track which individual cases gain or lose a passing \emph{written answer} across languages and use case-level scoring to identify a legal condition one answer overlooked.

\textbf{Scoring open-ended advice.}  ShARC asks a clarifying question when a rule requires a missing fact \citep{sharc}, and ConditionalQA associates answers with the conditions under which they apply \citep{sun}. ContractNLI identifies the contract passages that establish or contradict a proposition \citep{contractnli}. FActScore decomposes a generated text into factual claims \citep{factscore}, ALCE evaluates the quality of its citations \citep{alce}, and RAGTruth localizes unsupported text in retrieval-augmented answers \citep{ragtruth}. Our rubric also checks for omissions. An answer may contain only true statements and still leave out a condition the worker needs.

\textbf{Evidence sufficiency.} Changing a passage can change the answer, its evidentiary basis, or both. Substituting factual entities can put a passage in conflict with a model's prior knowledge \citep{longpre}. Moving an otherwise identical passage within a longer context can also change answer accuracy \citep{liu}. \citet{joren} distinguish whether the context suffices from whether the model answers correctly. We remove a designated passage while retaining the same legal answer key. This makes it possible to observe a particular legal conclusion surviving the loss of its supporting authority.

%% file: tables/archive_causes.tex
{\setlength{\tabcolsep}{3.8pt}\begin{tabular}{@{}lrrrrrrrrr@{}}
\toprule
 & \multicolumn{2}{c}{Strict pass} & \multicolumn{7}{c}{Failed positions, by first applicable reason} \\
\cmidrule(lr){2-3}\cmidrule(l){4-10}
Model & Rate (\%) & Count & \shortstack[r]{Unresolved\\transport} & \shortstack[r]{Used\\tool} & \shortstack[r]{Empty\\at cap} & \shortstack[r]{Text\\fragment} & \shortstack[r]{Legal\\error} & \shortstack[r]{Missing\\point} & \shortstack[r]{Wrong\\language} \\
\midrule
GPT-6.1 Sol & 76.2 & 1,463 & 0 & 44 & 0 & 0 & 276 & 118 & 19 \\
GPT-6 Luna & 48.9 & 938 & 0 & 0 & 0 & 0 & 654 & 305 & 23 \\
Claude Sonnet 5.5 & 44.5 & 854 & 0 & 0 & 31 & 0 & 876 & 131 & 28 \\
Kimi K3.1 Pro & 40.6 & 780 & 0 & 0 & 338 & 1 & 631 & 163 & 7 \\
Gemini 3.5 Flash-Lite & 25.5 & 489 & 0 & 0 & 0 & 5 & 1,166 & 258 & 2 \\
Gemma 4 31B IT & 21.1 & 405 & 0 & 0 & 358 & 0 & 877 & 280 & 0 \\
GLM-5.3 & 17.3 & 333 & 0 & 0 & 66 & 1 & 1,370 & 141 & 9 \\
Gemma 4 26B A4B IT & 14.6 & 280 & 0 & 0 & 1,005 & 0 & 428 & 198 & 9 \\
DeepSeek V4.1 Flash & 13.8 & 265 & 0 & 0 & 1,558 & 0 & 61 & 30 & 6 \\
\bottomrule
\end{tabular}}

%% file: deepseek_budget_note.tex
An eight-question pilot changed only the requested token cap at the endpoint's default reasoning setting. All eight responses exhausted 2,048 tokens without answer text; all eight returned complete text at 32,768. In a larger historical run, 1,280 of 1,365 identical prompts that exhausted the formal 2,080-token cap without an answer yielded text at 32,768 (93.8\%). The median long-budget answer on matched prompts used 9,565 completion tokens, with a median reasoning count of 8,744.5 across those answers. The larger comparison also changed the system instruction and reasoning setting. Both comparisons measure delivery; comparable legal scores are unavailable for the historical answers.

%% file: tables/pair.tex
{\setlength{\tabcolsep}{3.8pt}\begin{tabular}{@{}lrrrrrr@{}}
\toprule
Model & Pairs & Lost & Gained & \shortstack[r]{Any change\\\% [95\% CI]} & \shortstack[r]{Net change\\pp [95\% CI]} & \shortstack[r]{Loss among\\original passes (\%)} \\
\midrule
GPT-6.1 Sol & 1680 & 139 & 114 & 15.1 [12.0, 18.3] & -1.49 [-5.12, +2.26] & 10.7 \\
GPT-6 Luna & 1680 & 156 & 158 & 18.7 [15.6, 21.7] & +0.12 [-3.69, +3.99] & 19.0 \\
Claude Sonnet 5.5 & 1680 & 199 & 125 & 19.3 [16.1, 22.6] & -4.40 [-8.51, -0.42] & 24.5 \\
Kimi K3.1 Pro & 1680 & 150 & 162 & 18.6 [15.4, 21.9] & +0.71 [-3.33, +4.76] & 22.3 \\
Gemini 3.5 Flash-Lite & 1680 & 135 & 120 & 15.2 [12.0, 18.5] & -0.89 [-4.64, +2.80] & 30.6 \\
Gemma 4 31B IT & 1680 & 107 & 96 & 12.1 [9.1, 15.2] & -0.65 [-4.23, +2.68] & 29.4 \\
GLM-5.3 & 1680 & 134 & 107 & 14.3 [11.4, 17.5] & -1.61 [-5.36, +1.96] & 42.5 \\
Gemma 4 26B A4B IT & 1680 & 85 & 69 & 9.2 [6.5, 12.0] & -0.95 [-3.93, +1.96] & 32.8 \\
DeepSeek V4.1 Flash & 1680 & 73 & 74 & 8.8 [6.3, 11.4] & +0.06 [-2.80, +2.74] & 31.6 \\
\bottomrule
\end{tabular}}

%% file: tables/archive_language_counts.tex
\begin{tabular}{@{}lrrrrrrrrr@{}}\toprule
 & \multicolumn{8}{c}{Question and required answer language} &  \\
\cmidrule(lr){2-9}Model & Arabic & German & English & French & Japanese & Dutch & Port. & Chinese & \shortstack{Relative\\drop (\%)} \\
\midrule
GPT-6.1 Sol & 186 & 179 & 187 & 180 & 184 & \underline{178} & 180 & \textbf{189} & 5.8 \\
GPT-6 Luna & 120 & 119 & \underline{105} & 114 & 120 & 118 & \textbf{123} & 119 & 14.6 \\
Claude Sonnet 5.5 & 106 & 112 & \underline{90} & 107 & 103 & 106 & 113 & \textbf{117} & 23.1 \\
Kimi K3.1 Pro & \underline{89} & 93 & \textbf{112} & 95 & 92 & 103 & 101 & 95 & 20.5 \\
Gemini 3.5 Flash-Lite & \underline{57} & \textbf{65} & 61 & 63 & 62 & 61 & 58 & 62 & 12.3 \\
Gemma 4 31B IT & 46 & 50 & 57 & 46 & 53 & \underline{44} & 51 & \textbf{58} & 24.1 \\
GLM-5.3 & 39 & \textbf{46} & 45 & 40 & \underline{37} & \textbf{46} & 41 & 39 & 19.6 \\
Gemma 4 26B A4B IT & \underline{33} & 37 & \underline{33} & \textbf{38} & 34 & \underline{33} & \textbf{38} & 34 & 13.2 \\
DeepSeek V4.1 Flash & \textbf{45} & 39 & 34 & 30 & 27 & 30 & 34 & \underline{26} & 42.2 \\
\bottomrule\end{tabular}

%% file: tables/country_heatmap.tex
\includegraphics[width=\linewidth]{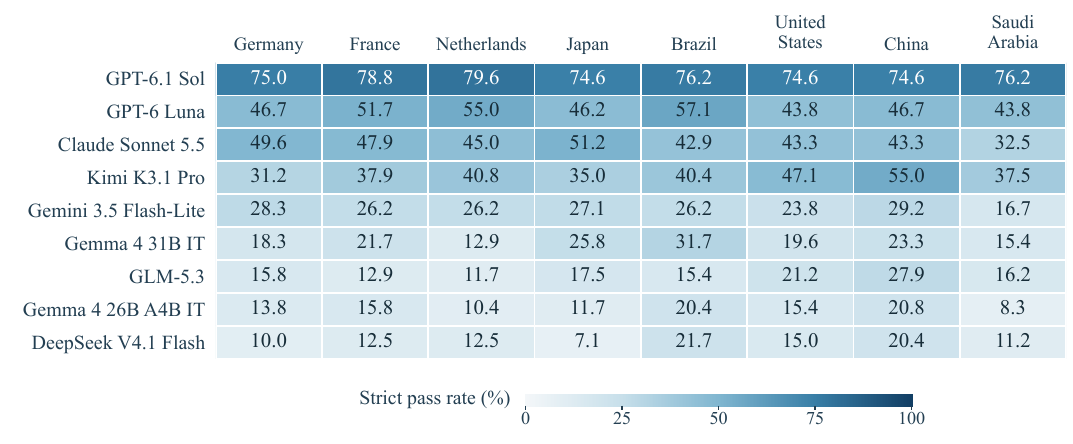}

%% file: evidence.tex
\section{Evidence Ablation Study}\label{sec:evidence}

\subsection{Diagnostic sample and evidence conditions} The evidence study asks how the same model answers the same legal question as its sources change. Sol, Sonnet, Kimi and DeepSeek each receive all 128 prompts in four conditions. Each jurisdiction contributes one direct question and one scenario. The 16 families comprise five Easy, six Medium and five Hard problems across seven topics. None has dismissal and termination as its primary topic. Families were selected before outcomes were known, subject to date, format and tier criteria, with a deterministic hash used to break ties. DeepSeek joined the roster after a 16-prompt pilot. Its 48 matching E1--E3 responses count once.

\begin{description}[leftmargin=3.7em,labelwidth=2.6em,labelsep=.55em,itemsep=1pt,topsep=3pt,parsep=0pt]

\item[E0] \emph{Closed book.} Reuse the main-study response to the same question in the same language.

\item[E1] \emph{Governing evidence.} Supply target-jurisdiction passages that together support every required proposition. The instructions ask the model to use the packet, identify gaps and distinguish reliance on prior knowledge.

\item[E2] \emph{Additional jurisdictions.} Interleave the E1 material with one original-language passage from each of the seven other jurisdictions.

\item[E3] \emph{Governing evidence removed.} Delete one prespecified target unit from E2, preserving the remaining text and order. Four families lose all target units and twelve retain some.

\end{description}

The seven foreign passages vary in how closely they match the question. Across the 16 families, E2 adds 112 placements (65 distinct foreign excerpts) to the 54 target units. Reviewers classified 52 placements as matching the subtopic, 31 as matching a related legal operation and 29 as weak matches. All seven matches are weak for both Shanghai heat protection and the French employee election.

Removing one governing passage leaves a known gap in every E3 packet. Of the 61 required propositions across the 16 families, 28 retain full support, 14 retain partial support and 19 lose it entirely. The legal scoring key stays fixed across conditions, so an answer can still earn credit for a correct conclusion after its supporting source disappears. We verified the frozen question, ordered passages and instructions against the actual inputs at all 2,048 positions.

\subsection{Key observations on strict passes} 
\begin{figure}[!htbp]\centering\includegraphics[width=\linewidth]{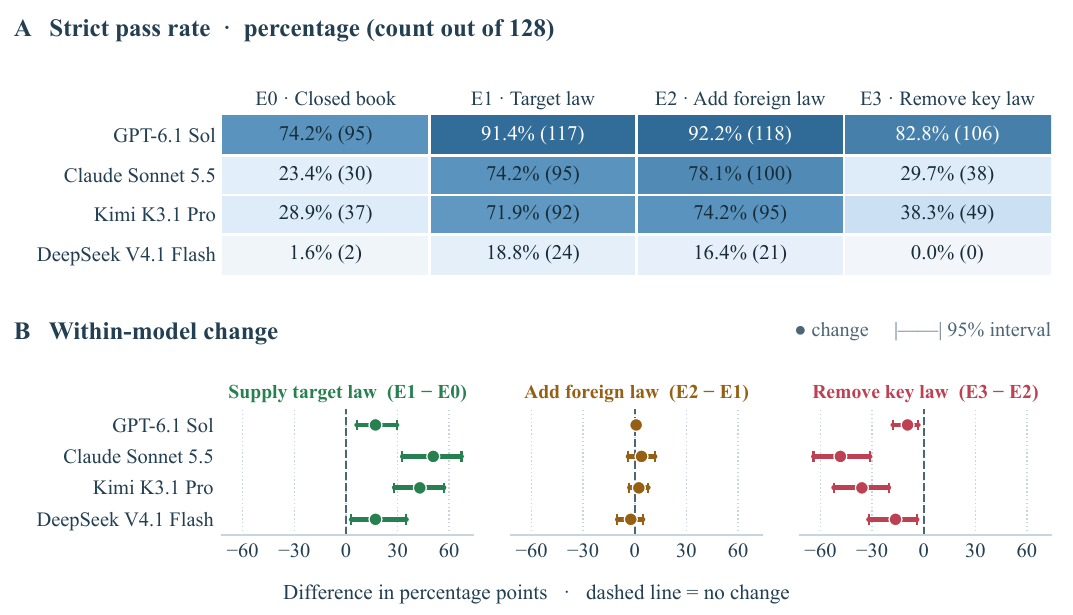}\caption{Panel A reports the strict pass rate and count on the same 128 prompts per model. In panel B each dot is the later rate minus the earlier rate for that model. The dashed zero line means equal pass rates. For example, GPT-6.1 Sol rises from 74.2\% at E0 to 91.4\% at E1, giving $+17.2$ percentage points. Bars are 95\% intervals from resampling 16 whole families.}\label{fig:evidencepaired}\end{figure}

\textbf{Passes rise with governing law and fall when a rule is removed.} Providing the governing passages and evidence-use instructions increases strict pass rates for all four models on the same 128 prompts (Figure~\ref{fig:evidencepaired}).

\textbf{Adding foreign passages changes the average little.} Each model’s interval includes zero. The passages give source cues and vary in relevance, so this comparison establishes little about unlabeled or more closely matched distractors. Removing the designated rule then lowers strict pass rates for all four models. Appendix Table~\ref{tab:evidencecontrast} gives the paired intervals.

\textbf{Supplying the law can close a performance gap.} Sonnet with governing evidence matches the average performance of closed-book Sol on these questions (74.2\% strict passes for each). With the same evidence, Kimi reaches 71.9\%, DeepSeek 18.8\% and Sol 91.4\%. Table~\ref{tab:evidencearm} shows the overlap underlying those averages. As one example, on California paid sick leave, Sonnet with evidence passes all eight language versions and closed-book Sol passes two. Both English responses calculate the same accrual, permitted use and carryover. Sonnet also checks that the worker qualifies to use the leave after 90 days, a required point omitted by Sol. Estimating the gain in a deployed system would require measuring how often its retriever supplies the governing provisions.

\begin{table}[!htbp]\centering\small\setlength{\tabcolsep}{3.4pt}
\begin{tabular}{lrrrrrr}
\toprule
Model & \shortstack{E1\\passes} & \shortstack{Both\\pass} & \shortstack{E1 model\\alone} & \shortstack{Sol E0\\alone} & \shortstack{Neither\\passes} & \shortstack{Net change (pp)\\{[95\% interval]}} \\
\midrule
Claude Sonnet 5.5 & 95 & 76 & 19 & 19 & 14 & 0.0 [-12.5, +14.1] \\
Kimi K3.1 Pro & 92 & 74 & 18 & 21 & 15 & -2.3 [-18.8, +12.5] \\
DeepSeek V4.1 Flash & 24 & 22 & 2 & 73 & 31 & -55.5 [-72.7, -36.7] \\
\bottomrule\end{tabular}
\caption{Matched comparison with closed-book GPT-6.1 Sol (95/128 strict passes). The four overlap columns partition the same 128 prompts in every row. Net change is the E1 model’s rate minus Sol’s E0 rate. Family-bootstrap intervals are descriptive.}\label{tab:evidencearm}
\end{table}

\subsection{Key observations on failure recovery}

\textbf{Sonnet corrects legal errors. DeepSeek shortens its reasoning.} Among the responses that first pass with governing evidence, 49 of 70 Sonnet answers previously gave incorrect advice. For DeepSeek, 21 of 22 previously exhausted the allowance, and all 21 use fewer reasoning tokens at E1. Table~\ref{tab:evidencetransition} counts both directions for the same prompts. Table~\ref{tab:evidencerecovery} traces each new pass to the failure recorded on its closed-book attempt.

\begin{table}[!htbp]\centering\small\setlength{\tabcolsep}{3.4pt}
\begin{tabular}{lrrrr}
\toprule
Model & \shortstack{Failed E0\\failed E1} & \shortstack{Failed E0\\passed E1} & \shortstack{Passed E0\\failed E1} & \shortstack{Passed E0\\passed E1} \\
\midrule
GPT-6.1 Sol & 9 & 24 & 2 & 93 \\
Claude Sonnet 5.5 & 28 & 70 & 5 & 25 \\
Kimi K3.1 Pro & 35 & 56 & 1 & 36 \\
DeepSeek V4.1 Flash & 104 & 22 & 0 & 2 \\
\bottomrule\end{tabular}
\caption{A pair consists of one closed-book response (E0) and one response to the same prompt with sufficient governing evidence and evidence-use instructions (E1). Each row partitions all 128 prompts for one model.}\label{tab:evidencetransition}
\end{table}

\begin{table}[!htbp]\centering\small\setlength{\tabcolsep}{3.4pt}
\begin{tabular}{lrrrrrr}
\toprule
Model & \shortstack{New\\passes} & \shortstack{No text\\at cap} & \shortstack{Used\\search} & \shortstack{Legal\\error} & \shortstack{Missing\\point} & \shortstack{Wrong\\language} \\
\midrule
GPT-6.1 Sol & 24 & 0 & 5 & 9 & 9 & 1 \\
Claude Sonnet 5.5 & 70 & 1 & 0 & 49 & 20 & 0 \\
Kimi K3.1 Pro & 56 & 16 & 0 & 28 & 12 & 0 \\
DeepSeek V4.1 Flash & 22 & 21 & 0 & 1 & 0 & 0 \\
\bottomrule\end{tabular}
\caption{Why did E0 fail on questions that passed at E1? New passes are the failed-E0, passed-E1 column of Table~\ref{tab:evidencetransition}. The five reason counts sum to new passes. The closed-book response exhausted its budget without text, performed a prohibited search, made a material legal error, gave legally correct advice while omitting a required point, or gave correct and complete advice in the wrong language. The first applicable reason is counted.}\label{tab:evidencerecovery}
\end{table}

\textbf{How much reasoning did DeepSeek save?} On these 21 recovered prompts the closed-book response spent all 2,080 output tokens on reasoning. With the passages and evidence-use instructions, median reasoning use falls to 984 tokens (range 114 to 1,691) and median total output to 1,531 (208 to 2,080). Eighteen of the 21 use fewer total tokens. Across all 128 pairs, reasoning use falls on 44, stays equal on 82 and rises on two; the overall median remains 2,080. DeepSeek delivers substantive answers on 8 prompts at E0 and 44 at E1. One additional E1 response contains only a heading.

\textbf{Budget exhaustion.} Across the four conditions, 418 responses finish at the 2,080-token cap with no visible answer. The provider attributes all 2,080 tokens to reasoning on 417 of them and 2,079 tokens on the remaining response. At E3 only three prompts yield substantive answers, all truncated. The other 125 count as end-to-end failures at this budget. Their empty text gives no basis for scoring a legal conclusion.

\textbf{Failures after rule removal.} Of the 12 Sol E2 passes lost at E3, five involve a search violation and seven omit necessary points from otherwise correct advice. All 123 admissible Sol E3 answers are free of material legal error. Sonnet loses 67 passes and gains five. Its 67 losses include 53 materially wrong answers and 14 incomplete ones. All 21 DeepSeek E2 passes exhaust the E3 budget before returning answer text.

\textbf{Some cases resist even complete evidence.} The Dutch railway medical-examination family yielded zero strict passes for every model even with full evidence. On the English E1 question, Sol distinguishes hiring consent from railway fitness certification but omits checking existing qualifications and explicitly making assignment conditional on them. Kimi likewise omits those steps. DeepSeek’s visible response ends before the statutory credential workflow. On the Shanghai heat case, deleting the designated rule reduces Sonnet’s eight passes to zero, while Kimi retains all eight. Appendix~\ref{app:evidencefamily} gives the complete family breakdown.

\subsection{Legal correctness, source support and gap recognition}

\textbf{Correctness and support can diverge.} A model may recall a correct rule after its source has been removed. We therefore score whether the supplied passages fully support each covered proposition and whether the model explicitly identifies the E3 gap. A partially supported proposition earns no full grounding credit. Correct and complete legal content requires every proposition and no material error, independent of answer language. Empty responses and searches remain in the end-to-end denominator and have no grounding or gap-recognition score.

\begin{figure}[!htbp]\centering\includegraphics[width=\linewidth]{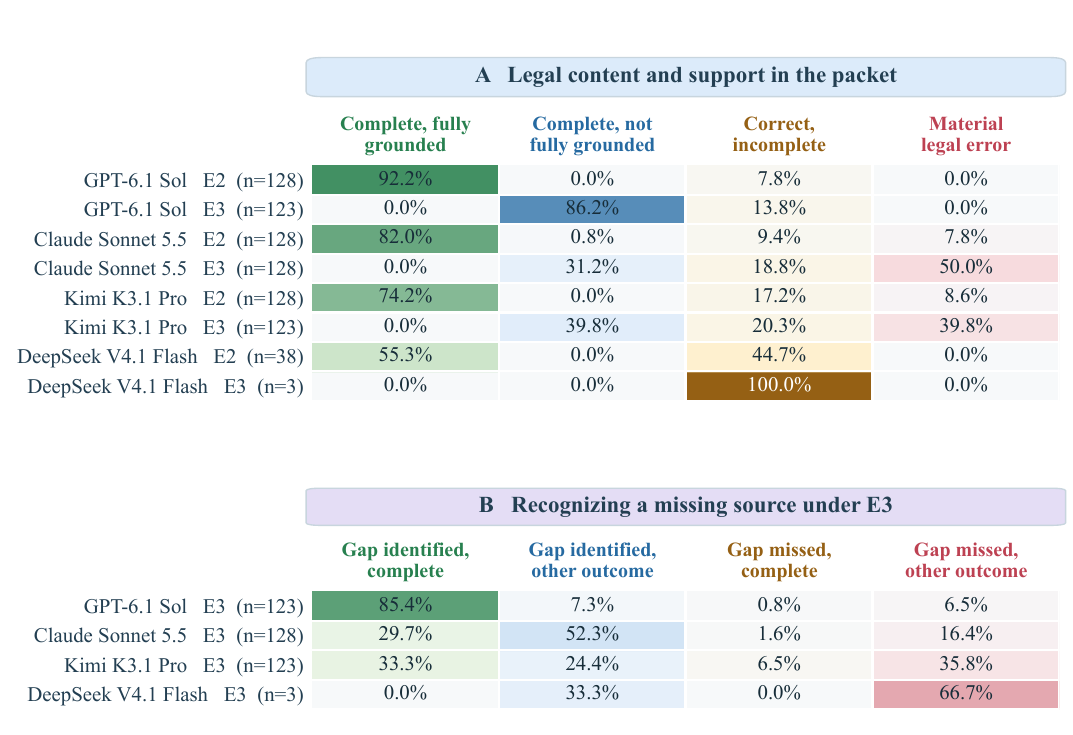}\caption{Percentages are conditional on the indicated number of eligible substantive answers ($n$). Each row sums to 100\% up to rounding. Complete requires every legal proposition and no material error. Fully grounded requires full packet support for all of those propositions. Every E3 packet lacks complete support for at least one required proposition by design.}\label{fig:evidencequality}\end{figure}

\textbf{A warning can accompany either a sound or a flawed answer.} Among E3 answers that identify the missing source, 105 of 114 from Sol give correct, complete advice, compared with 38 of 105 from Sonnet. Table~\ref{tab:evidencegap} counts these two judgments independently for E3. Figure~\ref{fig:evidencequality} also compares the support that eligible substantive answers receive from the E2 and E3 packets.

\begin{table}[!htbp]\centering\small\setlength{\tabcolsep}{3.4pt}
\begin{tabular}{lrrrrrr}
\toprule
Model & Scored & G+, C+ & G+, C− & G−, C+ & G−, C− & Grounded \\
\midrule
GPT-6.1 Sol & 123/128 & 105 & 9 & 1 & 8 & 211/471 \\
Claude Sonnet 5.5 & 128/128 & 38 & 67 & 2 & 21 & 209/488 \\
Kimi K3.1 Pro & 123/128 & 41 & 30 & 8 & 44 & 191/465 \\
DeepSeek V4.1 Flash & 3/128 & 0 & 1 & 0 & 2 & 0/13 \\
\bottomrule\end{tabular}
\caption{E3 gap recognition (G) crossed with legally correct, complete content (C) among judged answers. The first four counts sum to Scored. Grounded gives fully supported, covered propositions divided by all required propositions on grounding-scored answers. Appendix~\ref{app:evidencefamily} reports grounding within each cell.}\label{tab:evidencegap}
\end{table}

\begin{figure}[!htbp]\centering\includegraphics[width=\linewidth]{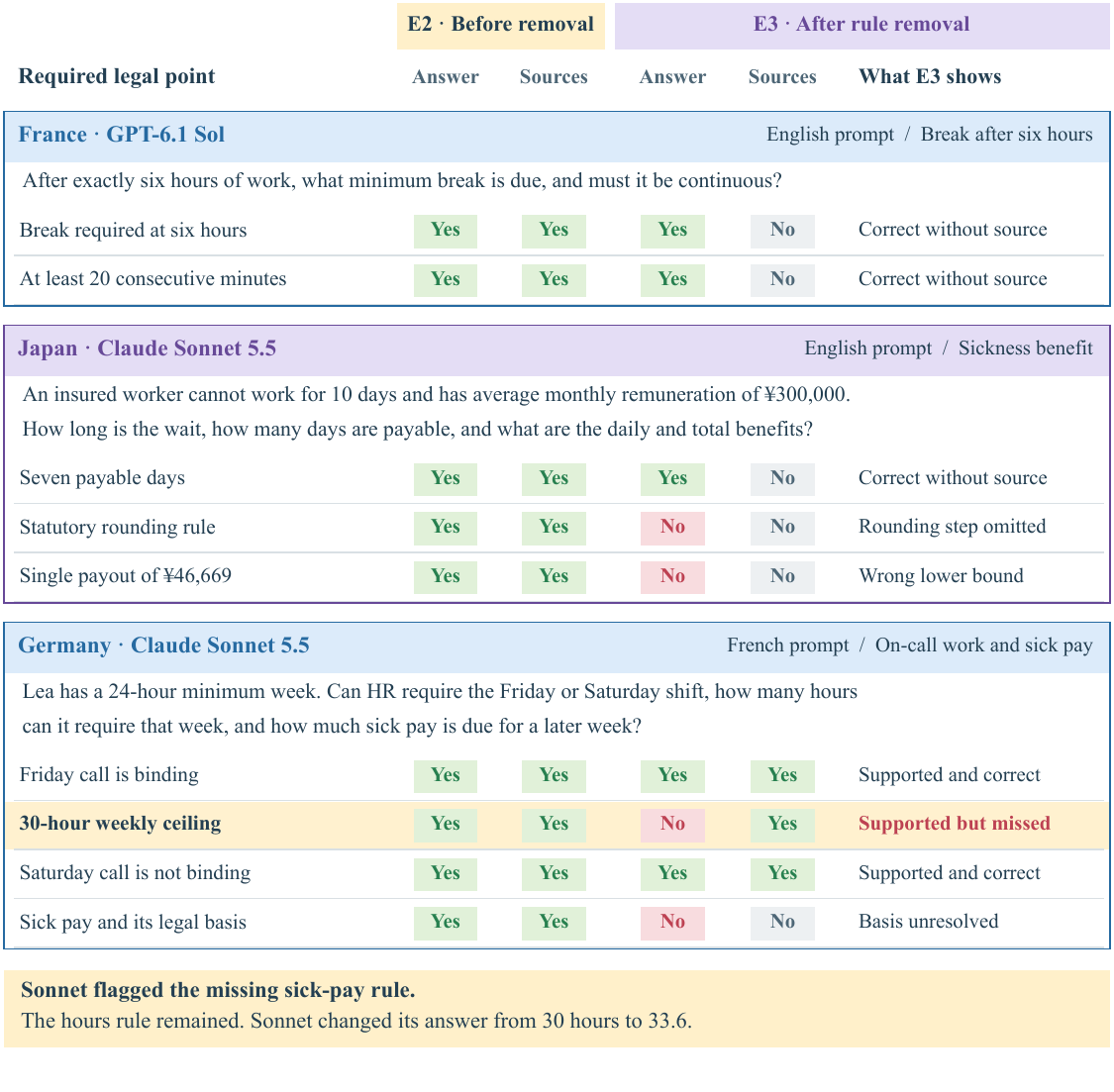}\caption{Each row follows one legal point across the same prompt before and after removal. \emph{Answer} means the response correctly and fully states that point. \emph{Sources} means the supplied passages suffice to establish it. E3 removes all French or Japanese authority in the first two cases. In the German case it removes only the sick-pay rule (TzBfG \S12(4)). The highlighted row identifies a fully supported point that Sonnet answered incorrectly despite recognizing the separate evidence gap.}\label{fig:pairedcases}\end{figure}

\Needspace{9\baselineskip}
\textbf{One missing rule leaves another decision fully answerable.} Figure~\ref{fig:pairedcases} distinguishes three outcomes after a rule is removed. Sol retains both French break requirements without their source. Sonnet retains the Japanese benefit duration but allows an incorrect lower payout. The German case tests what happens when the packet still establishes a decision. Only the sick-pay provision was removed. Sonnet explicitly identified that gap. It then accepted an agreement exceeding the separate statutory ceiling on hours, even though it had correctly applied the very same ceiling with the fuller packet. The Friday and Saturday scheduling decisions remained correct in both answers. Appendix~\ref{app:evidencecases} gives the underlying case and its scoring key.

%% file: archive_takeaways.tex
The following practices follow from the observed failure patterns. Their training effects remain to be tested on held-out families. Deployment recommendations concern the employment-law tasks and evidence conditions examined here.

\subsection{Evaluate models in the language in which advice must be delivered}

\textbf{Key takeaway.} Model choice depends on which language the user requires. On the same 240 problems, Kimi leads Sonnet in English (112 versus 90 strict passes), while Sonnet leads in Chinese (117 versus 95). Sonnet supplies correct and complete legal content more often in both languages, but 28 otherwise passing English responses use another language. These paired results show why legal quality and language compliance both matter to a multilingual workflow. They do not identify which languages were overrepresented in model training.

\textbf{Training recipe.} Construct multilingual examples that preserve the facts, governing jurisdiction, exceptions and required conclusions across versions. Audit the training mixture jointly by language and legal topic so that calculations and exceptions receive coverage in each language. Use explicit supervision for the requested answer language alongside proposition-level legal correctness. Evaluate training changes on held-out families, keeping all translations of a family in the same split. A balanced mixture is a hypothesis to test against the observed disparities.

\textbf{Deployment recipe.} Compare candidate models on a shared set of representative cases in the intended working language. Report strict success alongside correct-and-complete legal content, with failed deliveries retained in both denominators. Inspect cases where the two scores diverge before choosing a model or adding a translation stage. Any answer translation should be rechecked for altered thresholds, dates and qualifications. The experiment changes input and required output language together, so a workflow that changes only one requires its own evaluation.

\subsection{Measure whether evidence improves accuracy and reasoning efficiency}

\textbf{Key takeaway.} Evidence can change both the legal decision and the effort needed to produce it. On the 128 diagnostic prompts, Sonnet with governing passages reaches the same observed strict pass rate as closed-book Sol, 95/128. Of Sonnet's 70 newly passing responses, 49 previously made a material legal error. For DeepSeek, 21 of 22 new passes follow a closed-book attempt that exhausted its allowance without answer text. All 21 use fewer reasoning tokens with the passages and evidence-use instructions, with the median falling from 2,080 to 984. Across all 128 DeepSeek prompts the median remains 2,080. The gains describe a diagnostic sample and the combined change in evidence and instructions.

\textbf{Training recipe.} Pair tasks with sufficient evidence and tasks lacking a decisive provision, while marking which conclusions each packet supports. Reward correct application and completion of the required conclusions. Evaluate token-efficient training objectives at a fixed output allowance, checking for errors or omissions introduced by shorter answers and for unnecessary abstention on sufficient evidence. Budget-aware reasoning offers a relevant direction \citep{han2025budget}. When a decisive rule is absent, train the model to identify the specific missing authority and the decision that depends on it.

\textbf{Deployment recipe.} Evaluate the evidence packets the deployed retrieval process actually supplies. Track end-to-end correctness and completeness together with reasoning tokens, cap-exhausted empty answers and elapsed time. A model that answers accurately with verified governing passages may be suitable for a workload on which its closed-book result is weaker. That choice requires confirming source relevance, legal scope and version in the retrieved packets. The present evidence experiment supplies curated passages, so its gains do not establish the success rate of an automated retriever.

\subsection{Preserve supported conclusions when part of the evidence is missing}

\textbf{Key takeaway.} Identifying a missing authority is only one obligation in a partially supported task. In Figure~\ref{fig:pairedcases}, Sonnet correctly identifies the missing German sick-pay provision, while the packet retains the rules needed to establish the weekly hours ceiling. It nevertheless permits 33.6 hours instead of the supported 30-hour limit that it applied correctly at E2. This observed error concerns an answerable part of the task. The other two cases show why a correct conclusion can also survive without support from the supplied packet.

\textbf{Training recipe.} Start with a case containing several legal determinations and remove the authority for one while preserving complete support for another. Annotate support and correctness separately for each required proposition. Reward completion of the supported inference, identification of the unresolved decision, and a precise account of the authority or fact needed to resolve it. Preserve acceptable conditional branches when the facts warrant them. This extends evidence-sufficiency evaluation \citep{joren} to the individual conclusions within a single answer.

\textbf{Deployment recipe.} Require the response to connect each determination to the relevant available authority and identify any reliance on prior knowledge. Check whether the cited passage actually supports the conclusion and whether the answer omits a conclusion the packet establishes. A warning about missing evidence leaves those checks necessary. Under our fixed legal rubric, acknowledging the gap earns recognition credit but does not automatically satisfy a required legal conclusion. In an operational workflow, an unresolved decision can instead trigger retrieval of the missing provision or review by a qualified legal professional.

%% file: archive_conclusion.tex
GlobalHRBench shows that working language can change the success of employment-law assistance even when the legal problem stays fixed. Across eight jurisdictions, eight languages and nine models, GPT-6.1 Sol leads overall, while other models exhibit larger language gaps and ranking reversals. Separating legal content from answer-language compliance makes those reversals interpretable and relevant to model selection.

The evidence study shows why improving performance requires diagnosing the failure. Governing passages can repair legal errors and reduce the reasoning needed to deliver a correct answer. A model can also recognize missing authority while misapplying a rule the packet still supports. These findings motivate training that combines multilingual consistency, reasoning efficiency and proposition-level evidence use. For deployment, they support evaluating models in the user's working language and checking both the conclusions a file establishes and the decisions it leaves unresolved.

%% file: archive_limitations.tex
\textbf{Coverage and scale.} GlobalHRBench contains 240 question families with 1,920 language versions, evaluated across nine models. This provides systematic coverage of the selected jurisdictions and employment-law topics, while leaving room for a broader sample of legal decisions. Further funding would support additional original families, jurisdictions, languages and model versions, together with the legal and translation review needed to preserve alignment. Expanding the number of independent families would enable more detailed comparisons within each topic, jurisdiction and difficulty level.

\textbf{Sampling and diagnostic scope.} The balanced topic framework covers selected HR decisions, and the Hard tier is enriched for observed model errors. Estimating the frequency of these decisions in everyday practice would require sampling workplace requests. The evidence study uses 16 families and curated passages, with distractors that vary in relevance. A larger diagnostic sample and evaluation of retrieved evidence would help establish how its findings transfer to deployed systems. These extensions require resources for source verification, model execution and answer adjudication.

%% file: audit_appendix.tex
\section{Resource, execution and scoring audit}\label{app:audit}
\paragraph{Selection and review of the questions.}
Table~\ref{tab:allocationlaw} shows which decisions each topic covers and how the topics contribute to the three difficulty tiers. Replacing 19 Medium and 15 Easy families during selection reduced the sum of absolute deviations from the draft targets across the 64 cells from 62 to 24. The Hard set stayed fixed. T1--T8 follow the order in Figure~\ref{fig:allocation}.
\begin{table}[htbp]\centering\small\input{tables/archive_topic_tier}
\caption{Examples of employment decisions and actual case counts by difficulty. Each column contains 80 distinct families. The eight translations of an item count as one family.}\label{tab:allocationlaw}\end{table}

Historical screening includes both native and translated first responses, plus some shared-session records. In the selected Hard eighty, nine admission records explicitly designate a substantive failure on a non-original-language answer. This is different from discovering a defect in a translation, which should trigger correction of the item. Development and final evaluation overlap in the Sol family. The admission records name Sol for 78 and Astra for two. Fifteen admissions carry shared-session provenance. For those nine qualifying families, all nine original-language screening answers passed and 15 of 38 translated answers failed. One multilingual development log records 950 answers and 920 reviews across 123 staged families. A complete, deduplicated lifetime total across all candidate revisions cannot be identified from this log. No selected family records a defective translation itself as the qualifying error. These pathways qualify claims about difficulty in fresh independent sessions.

Frozen records preserve original-language authority, applicable scope, scenarios, gold answers, translations, scored propositions and human review flags. A conditional answer can be required when a fact is deliberately unspecified. The definitive bank includes 167 revised prompts in three sets of 95, 13 and 59. The final set comprises 52 jurisdiction clarifications, six localized material labels and one restored fee direction and statutory limit. Reuse checks compare actual prompt and instruction bytes within the model route, and apply the effective scoring overlay. The current repository contains version-specific approvals as well as older snapshot flags. Selection, review and authorization refer to different stages.

\paragraph{Requests and first attempts.} The unchanged (15,716), revised-prompt (1,503) and historical-exception (61) positions partition the 17,280 main positions. Revised prompts use the first identifiable model response under the corrected input. The 512 ablation E0 positions consist of 456 legacy reuse positions, 52 shared revised requests and four historical Sol positions. The audit traces all 512 to the same selected dispatch as the corresponding main positions. For the four historical Sol E0 positions, the earlier search-bearing answer is the formal attempt. The audit also restored two original evidence responses and scored them after identifying later responses on identical inputs. A returned completed search is disqualifying, even when followed by a valid-looking body. Tool use is established from returned events and annotations. The main table covers all 1,920 prompts per model.

\paragraph{Technical recovery.} Fifteen original dispatches lacked an identifiable model result: Sol had ten timeouts and two HTTP 500 responses, Gemini had two HTTP 502 responses, and Gemma 31B received an HTTP 200 response containing only whitespace. Each position had exactly one later dispatch using the identical payload and route. Fourteen of these dispatches were already archived; the remaining Portuguese question on Japanese law was transmitted once during the audit. Fourteen returned eligible answers (eight strict passes, six content failures). The fifteenth returned text and a recorded web search, a protocol failure. Two independent readings of the new Portuguese answer agreed that it covered two of six required points. No evidence-condition position changed. The historical sensitivity instead retains the original technical outcomes.

\paragraph{Counting requests.} The 19,328 main and evidence-condition positions use 18,816 distinct dispatches and 18,816 distinct provider response IDs. The full run ledger contains 20,695 dispatches, including replaced prompts, supplemental results and technical attempts. Identical request payloads may correspond to separate billable dispatches.

\paragraph{Runtime configuration.} Requests in the final ledger are dated 21--22 November 2026 UTC. Sol used the direct VexCloud route with a requested maximum of 2,048 output tokens, low effort, an empty tool list and tool selection disabled. DeepSeek V4.1 Flash used the Beijing Bailian endpoint with a 2,080-token limit, low effort and search disabled. The OpenRouter routes pinned the provider with no fallback, supplied no tools and requested 2,080 tokens. Sonnet 5.5 used Anthropic, Luna used OpenAI, Gemini 3.5 Flash-Lite used Google AI Studio, GLM-5.3 used Tenstorrent FP8, Kimi K3.1 Pro used Moonshot MXFP4 and both Gemma 4 models used DeepInfra FP8. Non-Gemma requests in those routes specified low effort. Gemma enabled reasoning without a named intensity. The dated Luna request returned the unversioned Luna alias. Temperature was omitted in the selected requests. Identically named options do not establish equal compute across endpoints.

Sol's direct instruction prohibited tools and search. The other main instruction additionally enumerated browsing, retrieval and external legal materials. Fifty-five Sol primary main receipts reported output tokens above the requested maximum, including 49 above 2,080. We retain the authorized, otherwise valid visible text and separately report a cap-exclusion sensitivity in the audit. Per-request finish reasons, metering, hashes, route, timestamps and technical attempts accompany the ledger. Reported charges omit unpriced direct-provider receipts.

\paragraph{Answer adjudication.} Initial local, model-assisted graders received blinded answer packets with reference conclusions and required points. A stratified pre-adjudication cross-check covered 420 outputs from an earlier main scoring stage. Initial strict-pass labels agreed on 396/420; 74 pairs differed under the wider comparison across grading dimensions. Those disagreements were adjudicated. Of these, 386 were retained in the later checkpoint. Matching the actual first response and effective scoring keys leaves 385 in the present analysis; their pre-adjudication strict agreement was 364/385 ($\kappa=0.886$). The later position queue represents 2,719 distinct response hashes, including 37 E0 bodies shared by the main and ablation queues. Its 2,756 position-level judgments comprise 1,152 revised-main, 1,592 ablation and 12 historical entries. A reused E0 answer can appear in multiple scoring processes. This count cannot serve as independent agreement or as evidence of specialist human validation.

The audit compared shared E0 answers across scorers. Twenty-four shared bodies had discrepancies across correctness, coverage or language labels, including nine strict-pass switches. Human-readable rationales and response-bound adjudications accompany the revised grades. One Brazilian benefits answer retains a scope-specific one-point coverage difference and fails both strict keys. A six-answer stratified spot review also exposed inconsistent application of a wage condition. All 72 positions in that California hiring-notice family were checked, including each of 63 visible answers. Twelve labels changed. The strict count moved from ten to eight: three earlier passes were rejected and one earlier failure was accepted. Where reviewer identity or version is unavailable, the audit identifies roles by the recorded label. It supplies no estimate of legal-expert agreement from those labels.
\paragraph{Interpreting answer counts.}
Table~\ref{tab:component} uses substantive, protocol-eligible answers for the last four counts. Some returned text contains only an opening fragment. Some visible Sol answers followed a search. Their text presence is recorded even when they receive no eligible legal grade.
\begin{table}[!htbp]\centering\small\input{tables/body}
\caption{Main-study counts. Visible text, content-gradable text and strict passes have different meanings. A required-point-complete answer can still make an additional material error.}\label{tab:component}\end{table}

Table~\ref{tab:tags} groups the substantive failures by the legal issue the reviewer identified.
\begin{table}[!htbp]\centering\small\input{tables/content_tags}
\caption{Reviewer-assigned error groups among substantive strict failures, one archived tag per answer. Other includes unclassified, missing, language, procedural and citation labels. This classification supplements the metric-derived partition in Table~\ref{tab:fail}.}\label{tab:tags}\end{table}

%% file: tables/archive_topic_tier.tex
\begin{tabularx}{\linewidth}{@{}>{\raggedright\arraybackslash}Xrrr@{}}
\toprule
HR decision and legal issues & Easy & Medium & Hard \\
\midrule
\textbf{T1 Contracts and employment.} \textit{Formation, probation and fixed-term renewal.} Decide whether a probation term is valid or renewal changes the contract. & 11 & 11 & 9 \\
\textbf{T2 Wages and pay.} \textit{Wage floors, pay dates and deductions.} Identify the rate, payer or due date and calculate lawful deductions. & 12 & 14 & 10 \\
\textbf{T3 Hours and overtime.} \textit{Working time, overtime limits, premiums and breaks.} Reconstruct a shift, test a rest interval or calculate overtime pay. & 14 & 9 & 9 \\
\textbf{T4 Leave and benefits.} \textit{Annual, sick, maternity and other family leave.} Determine eligibility, payable days and how entitlements interact. & 12 & 10 & 18 \\
\textbf{T5 Termination.} \textit{Grounds, notice, severance and remedies.} Assess a dismissal and determine the notice, payment or remedy due. & 16 & 17 & 9 \\
\textbf{T6 Equality and protected workers.} \textit{Discrimination, pregnancy and young workers.} Identify special protections affecting hiring, scheduling or dismissal. & 8 & 9 & 8 \\
\textbf{T7 Occupational safety and health.} \textit{Prevention and work-related injury.} Use risk or incident records to identify prevention, reporting or protection duties. & 5 & 7 & 8 \\
\textbf{T8 Disputes and enforcement.} \textit{Grievances, inspection, dispute resolution and liability.} Choose the competent body and deadline and determine the effect of a procedural error. & 2 & 3 & 9 \\
\midrule All topics & 80 & 80 & 80 \\
\bottomrule\end{tabularx}

%% file: tables/body.tex
{\setlength{\tabcolsep}{3.8pt}\begin{tabular}{@{}lrrrrrr@{}}
\toprule
Model & \shortstack[r]{Visible\\text} & Scorable & \shortstack[r]{Legally\\correct} & Complete & \shortstack[r]{Requested\\language} & \shortstack[r]{Strict\\pass} \\
\midrule
GPT-6.1 Sol & 1920 & 1876 & 1600 & 1483 & 1844 & 1463 \\
GPT-6 Luna & 1920 & 1920 & 1266 & 967 & 1869 & 938 \\
Claude Sonnet 5.5 & 1889 & 1889 & 1013 & 912 & 1829 & 854 \\
Kimi K3.1 Pro & 1582 & 1581 & 950 & 799 & 1570 & 780 \\
Gemini 3.5 Flash-Lite & 1920 & 1915 & 749 & 512 & 1903 & 489 \\
Gemma 4 31B IT & 1562 & 1562 & 685 & 416 & 1559 & 405 \\
GLM-5.3 & 1854 & 1853 & 483 & 388 & 1828 & 333 \\
Gemma 4 26B A4B IT & 915 & 915 & 487 & 296 & 898 & 280 \\
DeepSeek V4.1 Flash & 362 & 362 & 301 & 277 & 353 & 265 \\
\bottomrule
\end{tabular}}

%% file: tables/content_tags.tex
{\setlength{\tabcolsep}{3.8pt}\begin{tabular}{@{}lrrrrrrrr@{}}
\toprule
Model & Rule & Applic. & Time & Calc. & Omit & Cutoff & Mixed & Other \\
\midrule
GPT-6.1 Sol & 86 & 44 & 45 & 33 & 131 & 1 & 3 & 70 \\
GPT-6 Luna & 246 & 119 & 81 & 106 & 313 & 3 & 42 & 72 \\
Claude Sonnet 5.5 & 391 & 129 & 108 & 107 & 154 & 4 & 67 & 75 \\
Kimi K3.1 Pro & 222 & 60 & 73 & 65 & 260 & 40 & 37 & 44 \\
Gemini 3.5 Flash-Lite & 458 & 159 & 114 & 141 & 311 & 55 & 125 & 63 \\
Gemma 4 31B IT & 352 & 140 & 84 & 63 & 352 & 26 & 81 & 59 \\
GLM-5.3 & 703 & 186 & 127 & 108 & 143 & 0 & 179 & 74 \\
Gemma 4 26B A4B IT & 152 & 52 & 29 & 12 & 282 & 49 & 21 & 38 \\
DeepSeek V4.1 Flash & 20 & 0 & 3 & 2 & 56 & 10 & 0 & 6 \\
\bottomrule
\end{tabular}}

%% file: example.tex
\section{Evidence results by family and response}\label{app:evidencefamily}

Table~\ref{tab:evidencecontrast} reports uncertainty for each successive change to the evidence packet.

\begin{table}[!htbp]\centering\small\setlength{\tabcolsep}{3.4pt}
\begin{tabular}{lrrr}
\toprule
Model & E1 $-$ E0 & E2 $-$ E1 & E3 $-$ E2 \\
\midrule
GPT-6.1 Sol & +17.2 [+6.2, +29.7] & +0.8 [0.0, +2.3] & -9.4 [-18.0, -3.1] \\
Claude Sonnet 5.5 & +50.8 [+32.8, +67.2] & +3.9 [-3.9, +11.7] & -48.4 [-64.1, -31.2] \\
Kimi K3.1 Pro & +43.0 [+28.1, +57.0] & +2.3 [-3.1, +7.8] & -35.9 [-52.3, -20.3] \\
DeepSeek V4.1 Flash & +17.2 [+3.1, +35.2] & -2.3 [-10.2, +4.7] & -16.4 [-32.0, -3.9] \\
\bottomrule\end{tabular}
\caption{Paired strict-pass differences in percentage points with 95\% percentile intervals. Each draw resamples 16 whole families and retains all eight language pairs per family. All 128 pairs contribute per model and contrast (5,000 draws, seed 20261122).}\label{tab:evidencecontrast}
\end{table}

Table~\ref{tab:evidencefamilies} gives each family’s four-condition trajectory. Table~\ref{tab:evidencedeletion} pools the E2 and E3 counts within each removal group.

\begin{table}[!htbp]\centering\small\setlength{\tabcolsep}{3.4pt}
\begin{tabular}{llcrrrr}
\toprule
Law & Family & Del. & \shortstack{GPT-6.1\\Sol} & \shortstack{Sonnet\\5.5} & \shortstack{Kimi\\K3.1 Pro} & \shortstack{DeepSeek\\V4.1 Flash} \\
\midrule
DE & Fixed-term probation & All & 8/8/8/8 & 3/7/7/8 & 3/6/7/3 & 0/7/4/0 \\
DE & On-call work and sick pay & Part & 6/8/8/8 & 0/5/6/2 & 0/4/5/1 & 0/0/0/0 \\
BR & Psychosocial risks & Part & 8/8/8/8 & 5/8/7/3 & 3/8/8/7 & 0/0/0/0 \\
BR & Protective leave and benefits & Part & 8/8/8/6 & 5/7/6/4 & 2/7/7/4 & 0/0/0/0 \\
CN & Medical leave and dismissal & Part & 6/8/8/7 & 5/4/7/2 & 6/8/7/7 & 0/0/1/0 \\
CN & Shanghai heat allowance & Part & 6/7/8/7 & 0/6/8/0 & 7/8/8/8 & 0/0/0/0 \\
FR & Six-hour break threshold & All & 8/8/8/8 & 6/7/7/6 & 8/8/8/7 & 2/6/8/0 \\
FR & Employee election challenge & Part & 3/8/8/8 & 4/7/6/5 & 0/6/7/2 & 0/0/0/0 \\
JP & Older-worker benefit rates & Part & 7/8/8/8 & 0/7/5/0 & 0/7/7/0 & 0/0/0/0 \\
JP & Sickness benefit calculation & All & 7/6/6/2 & 2/3/5/0 & 0/0/0/0 & 0/0/0/0 \\
NL & Railway medical examinations & Part & 0/0/0/0 & 0/0/0/0 & 0/0/0/0 & 0/0/0/0 \\
NL & Predictable work request & Part & 5/8/8/8 & 0/6/8/0 & 0/5/7/0 & 0/0/0/0 \\
SA & Unwritten employment contract & All & 8/8/8/5 & 0/7/7/0 & 6/8/8/1 & 0/8/5/0 \\
SA & Annual and examination leave & Part & 5/8/8/8 & 0/6/6/0 & 1/4/4/0 & 0/0/0/0 \\
US & Tips and commission exemption & Part & 8/8/8/8 & 0/7/7/3 & 1/6/4/1 & 0/1/2/0 \\
US & California paid sick leave & Part & 2/8/8/7 & 0/8/8/5 & 0/7/8/8 & 0/2/1/0 \\
\bottomrule\end{tabular}
\caption{Strict passes out of eight languages in E0/E1/E2/E3 order for every family. All indicates removal of the sole target-jurisdiction unit, Part removal of one among several. The seven foreign passages remain in both groups.}\label{tab:evidencefamilies}
\end{table}

\begin{table}[!htbp]\centering\small\setlength{\tabcolsep}{3.4pt}
\begin{tabular}{lrrrr}
\toprule
Model & Sole unit E2 & Sole unit E3 & Partial E2 & Partial E3 \\
\midrule
GPT-6.1 Sol & 30/32 & 23/32 & 88/96 & 83/96 \\
Claude Sonnet 5.5 & 26/32 & 14/32 & 74/96 & 24/96 \\
Kimi K3.1 Pro & 23/32 & 11/32 & 72/96 & 38/96 \\
DeepSeek V4.1 Flash & 17/32 & 0/32 & 4/96 & 0/96 \\
\bottomrule\end{tabular}
\caption{Strict passes by target-evidence removal group. All four sole-unit families are Easy. Each subgroup fixes the identity of its families across conditions. Differences between the two subgroups also reflect family composition.}\label{tab:evidencedeletion}
\end{table}

Table~\ref{tab:evidenceground} gives point denominators for the four E3 response cells. Table~\ref{tab:evidencebodies} shows which planned responses contribute a body. Table~\ref{tab:evidencequality} gives conditional content and grounding credit for all three evidence arms.

\begin{table}[!htbp]\centering\small\setlength{\tabcolsep}{3.4pt}
\begin{tabular}{lrrrr}
\toprule
Model & G+, C+ & G+, C− & G−, C+ & G−, C− \\
\midrule
GPT-6.1 Sol & 181/399 & 6/29 & 0/3 & 24/40 \\
Claude Sonnet 5.5 & 40/128 & 112/253 & 6/8 & 51/99 \\
Kimi K3.1 Pro & 65/141 & 23/97 & 13/31 & 90/196 \\
DeepSeek V4.1 Flash & --- & 0/3 & --- & 0/10 \\
\bottomrule\end{tabular}
\caption{E3 grounded proposition credit and possible points within the four cells of Table~\ref{tab:evidencegap}. A dash indicates a cell with zero answers. A source gap may leave several other propositions fully supported.}\label{tab:evidenceground}
\end{table}

Five original California scores awarded a full grounded point to a compound proposition whose accrual calculation and eligibility clauses remain supported but whose annual-use limitation depends on deleted Labor Code §246(d),(j). The frozen rubric labels this proposition partially supported and the other two unsupported. Applying the full-proposition rule removes one grounded point each from Sol (Chinese), Sonnet (English and Chinese), and Kimi (English and Chinese). The original E3 grounded totals were 212/471, 211/488, 193/465 and 0/13 in table order. Reconciliation changes only grounded credit. Both versions and their receipt identifiers are preserved.

\begin{table}[!htbp]\centering\small\setlength{\tabcolsep}{3.4pt}
\begin{tabular}{lrrrrr}
\toprule
Model & Condition & Text & Admissible scored & Substantive & Cap empty \\
\midrule
GPT-6.1 Sol & E0 & 128 & 122 & 122 & 0 \\
GPT-6.1 Sol & E1 & 128 & 128 & 128 & 0 \\
GPT-6.1 Sol & E2 & 128 & 128 & 128 & 0 \\
GPT-6.1 Sol & E3 & 128 & 123 & 123 & 0 \\
Claude Sonnet 5.5 & E0 & 127 & 127 & 127 & 1 \\
Claude Sonnet 5.5 & E1 & 128 & 128 & 128 & 0 \\
Claude Sonnet 5.5 & E2 & 128 & 128 & 128 & 0 \\
Claude Sonnet 5.5 & E3 & 128 & 128 & 128 & 0 \\
Kimi K3.1 Pro & E0 & 103 & 103 & 102 & 25 \\
Kimi K3.1 Pro & E1 & 128 & 128 & 128 & 0 \\
Kimi K3.1 Pro & E2 & 128 & 128 & 128 & 0 \\
Kimi K3.1 Pro & E3 & 123 & 123 & 123 & 5 \\
DeepSeek V4.1 Flash & E0 & 8 & 8 & 8 & 120 \\
DeepSeek V4.1 Flash & E1 & 45 & 45 & 44 & 83 \\
DeepSeek V4.1 Flash & E2 & 38 & 38 & 38 & 90 \\
DeepSeek V4.1 Flash & E3 & 3 & 3 & 3 & 125 \\
\bottomrule\end{tabular}
\caption{Response availability. Text includes search-bearing outputs. Admissible scored text includes two zero-scored truncations without a substantive legal answer (Kimi E0 and DeepSeek E1). Cap empty counts a returned receipt without answer text. Each row has 128 planned positions.}\label{tab:evidencebodies}
\end{table}

Table~\ref{tab:evidencequality} reports content quality, support and input length for the evidence-conditioned answers.

\begin{table}[!htbp]\centering\small\setlength{\tabcolsep}{3.4pt}
\begin{tabular}{lrrrrr}
\toprule
Model & Condition & Correct/text & Legal complete/text & Grounded points & Median input \\
\midrule
GPT-6.1 Sol & E1 & 128/128 & 118/128 & 475/488 & 1,207.5 \\
GPT-6.1 Sol & E2 & 128/128 & 118/128 & 475/488 & 3,622.5 \\
GPT-6.1 Sol & E3 & 123/123 & 106/123 & 211/471 & 3,276.5 \\
Claude Sonnet 5.5 & E1 & 115/128 & 103/128 & 467/488 & 1,995.5 \\
Claude Sonnet 5.5 & E2 & 118/128 & 106/128 & 463/488 & 5,448.5 \\
Claude Sonnet 5.5 & E3 & 64/128 & 40/128 & 209/488 & 5,072.5 \\
Kimi K3.1 Pro & E1 & 114/128 & 93/128 & 446/488 & 1,482.5 \\
Kimi K3.1 Pro & E2 & 117/128 & 95/128 & 454/488 & 4,172.0 \\
Kimi K3.1 Pro & E3 & 74/123 & 49/123 & 191/465 & 3,870.5 \\
DeepSeek V4.1 Flash & E1 & 44/45 & 27/45 & 102/133 & 1,334.5 \\
DeepSeek V4.1 Flash & E2 & 38/38 & 21/38 & 76/110 & 3,687.0 \\
DeepSeek V4.1 Flash & E3 & 3/3 & 0/3 & 0/13 & 3,432.0 \\
\bottomrule\end{tabular}
\caption{Evidence-conditioned quality among protocol-eligible, scored texts (including one zero-scored DeepSeek E1 heading). Correct means free of material error. Grounded compares fully supported, covered propositions to required propositions among grounding-scored texts. A local review of two Sol answers supplies grounding fields absent from their original scores (Dutch rail E1 in Arabic, 4/5; German on-call E2 in French, 4/4). Median input tokens use all 128 provider receipts per condition.}\label{tab:evidencequality}
\end{table}

\section{Two worked evidence interventions}\label{app:evidencecases}\small

These two preselected experimental families illustrate different conditional branches. The question text, English key and rubric below come from the actual frozen instruments. Quoted model text is taken from the English-question receipts. Markdown emphasis in quoted answers is typeset as emphasis. The accompanying machine-readable examples contain both language versions of every packet, complete cited answers, hashes and the ordered evidence texts.

\subsection{Which tips count in the commission denominator?}

\noindent\textbf{GHB-W2-US-14} Medium · QA · 2026-11-15.

\paragraph{Original English} {\emergencystretch=1.5em A qualifying retail restaurant uses a valid representative period for the federal commissioned-employee overtime exemption. The employee meets the regular-rate requirement in every overtime week. For the period the employee receives \$4,000 in direct noncommission wages, \$4,600 from service charges that qualify as bona fide commissions, and \$2,000 in additional voluntary tips. The restaurant actually and lawfully applies \$600 of the \$2,000 to a state minimum-wage obligation. It claims no federal tip credit. Under the current U.S. Department of Labor Wage and Hour Division interpretation, calculate the commission share and decide whether the commission condition is met. Recalculate if the employer instead took no tip credit under any law and the three dollar totals stayed the same. In that no-credit version, would the answer on the commission condition differ if the governing federal court follows the Fourth Circuit’s holding on restaurant tips? Do not decide state overtime liability.\par}

\paragraph{Frozen French translation} {\emergencystretch=1.5em Un restaurant de détail admissible utilise une période représentative valide pour l'exemption fédérale des heures supplémentaires applicable aux salariés rémunérés à la commission. Le salarié satisfait à la condition relative au taux normal de rémunération pour chaque semaine où il effectue des heures supplémentaires. Au cours de la période, il reçoit 4 000 dollars en salaire direct hors commissions, 4 600 dollars provenant de frais de service qui constituent de véritables commissions, et 2 000 dollars supplémentaires de pourboires volontaires. Le restaurant affecte effectivement et légalement 600 des 2 000 dollars de pourboires à la satisfaction d'une obligation de salaire minimum de l'État. Il n'utilise aucun crédit fédéral au titre des pourboires. Selon l'interprétation actuelle de la division des salaires et des heures de travail du ministère américain du Travail, calculez la part des commissions et indiquez si la condition relative aux commissions est satisfaite. Refaites le calcul si l'employeur n'utilise en fait aucun crédit au titre des pourboires sur le fondement d'aucune loi et que les trois montants restent inchangés. Dans cette version sans crédit, la réponse sur la condition des commissions changerait-elle si le tribunal fédéral compétent suivait la décision du quatrième circuit sur les pourboires dans la restauration ? Ne statuez pas sur l'obligation de payer les heures supplémentaires en vertu du droit de l'État.\par}

\paragraph{Gold answer.} WHD counts only tips actually used to satisfy a wage obligation in the denominator, including a state tip credit; bona fide voluntary tips are never commissions here. With \$600 of credits the share is \$4,600/\$9,200 or 50\% exactly, insufficient. With no credit the WHD share is \$4,600/\$8,600 or about 53.49\%, sufficient. The Fourth Circuit counts all tips as compensation under Wai Man Tom. In the no-credit version its denominator is \$10,600, about 43.40\%, insufficient.

\paragraph{Scored propositions.}\begin{enumerate}
\item WHD includes actual \$600 state tip credit and excludes unused voluntary tips; commission share 50\%, not enough
\item Without any credit WHD ratio about 53.49\%, satisfies more-than-half requirement
\item Fourth Circuit approach counts all \$2,000 tips as compensation and no voluntary tips as commissions; no-credit share about 43.40\%, fails
\end{enumerate}

\paragraph{Evidence and conditional branches.} The three E1 units give the strict majority threshold under \href{https://uscode.house.gov/us/usc/119-134/t29/s207/}{29 U.S.C. §207(i)}, the Wage and Hour Division (WHD) treatment of tips actually used for wage obligations in \href{https://www.dol.gov/sites/dolgov/files/WHD/opinion-letters/FLSA/FLSA2026-4.pdf}{Opinion Letter FLSA2026-4, Part B, p. 5}, and the Fourth Circuit treatment of tips in \emph{\href{https://www.ca4.uscourts.gov/opinions/182509.P.pdf}{Wai Man Tom v. Hospitality Ventures LLC}}, slip opinion pp. 19--20. The seven E2 passages concern overtime or working time under the other jurisdictions. E3 removes only the court opinion. The restaurant facts, the WHD letter and the statutory threshold still resolve the first two branches. The alternate judicial denominator loses its controlling passage. P1 and P2 retain full support in E3. P3 loses support. The judicial branch changes the authority governing the same no-credit facts. It keeps both the compensation inputs and the question about the federal exemption fixed.

\paragraph{Sonnet, E2.} “It rejected the view that tips not used to satisfy FLSA obligations may be left out.” The answer includes all tips, reports about 43.4\% in the judicial branch and receives all three content and grounding points.

\paragraph{Sonnet, E3.} “Following the Fourth Circuit, the no-credit answer does not differ.” The body flags the missing holding, then excludes tips and reports the wrong result. Its two WHD conclusions receive two content and two grounded points out of three.

\paragraph{Sol, E3.} “The supplied materials do not include the Fourth Circuit’s decision.” Sol obtains all three ratios and identifies the missing opinion, while attributing the rule to McFeeley instead of Wai Man Tom. The rubric awards three content points and two grounded points.

\subsection{Which collective-agreement terms change the call-out?}

\noindent\textbf{GHB-EX-NEW-DE-S14} Hard · SCENARIO · 2026-11-15.

\paragraph{Original German} {\emergencystretch=1.5em Ein deutscher Betrieb beschäftigt Lea auf Abruf. Vereinbart sind mindestens 24 Stunden pro Woche. Die festgelegten Referenztage sind Montag bis Freitag, jeweils 8–18 Uhr. Ein wirksam geltender Tarifvertrag regelt sowohl die tägliche und wöchentliche Arbeitszeit als auch die Ankündigungsfrist. Er lässt zweistündige Einsätze, Ankündigungen mit zwei vollen Tagen Vorlauf und zusätzliche Abrufe von bis zu 40 \% der Mindestwochenzeit zu. Die Regelungen sind voneinander unabhängig. In einer Woche hat Lea bis Donnerstag 28 Stunden geleistet. Am Montag wurde ihr ein weiterer Einsatz für Freitag von 12 bis 14 Uhr angekündigt. Als Alternative möchte HR dafür am Samstag von 12 bis 14 Uhr abrufen. Welche dieser Anordnungen ist verbindlich? Wie viele Stunden darf HR in dieser Woche insgesamt einseitig abrufen? In einer späteren Woche ist Lea durchgehend arbeitsunfähig; ihr gesetzlicher Entgeltfortzahlungsanspruch steht außer Streit. Im ganzen dreimonatigen Referenzzeitraum liegen genau 13 volle Wochen, davon zwei vollständige Urlaubswochen und eine volle Woche Kurzarbeit Null. In den übrigen zehn Wochen arbeitete Lea zusammen 294 Stunden ohne Überstunden oder Zuschläge. Sie verdient 20 Euro je Stunde. HR plant 24 Stunden Krankenlohn. Wie hoch ist die Zahlung für die volle Krankheitswoche? Es gibt keine günstigere Berechnungsregel.\par}

\pagebreak

\paragraph{Frozen English translation} {\emergencystretch=1.5em A German business employs Lea on an on-call basis. The agreed minimum is 24 hours per week. The designated reference days are Monday to Friday, from 08:00 to 18:00 each day. A valid and applicable collective agreement governs both daily and weekly working time and the notice period. It allows two-hour assignments, notice given two full days in advance, and additional call-outs of up to 40\% of the weekly minimum. These provisions operate independently. In one week, Lea has worked 28 hours by Thursday. On Monday she was notified of an extra assignment for Friday from 12:00 to 14:00. As an alternative, HR wants to call her in on Saturday from 12:00 to 14:00. Which of these instructions is binding? How many hours in total can HR require unilaterally in that week?

In a later week, Lea is unable to work throughout the week; her statutory entitlement to continued pay is undisputed. The full three-month reference period contains exactly 13 full weeks, including two full weeks of annual leave and one full week of short-time work reduced to zero hours (Kurzarbeit Null). In the other ten weeks, Lea worked 294 hours in total, without overtime or wage premiums. She earns €20 per hour. HR plans to pay sick pay for 24 hours. How much is payable for the full week of illness? There is no more favorable calculation rule.\par}

\paragraph{Gold answer.} Section 12(6) of the German Part-Time and Fixed-Term Employment Act (TzBfG) allows a collective agreement to derogate from subsection 1 for short daily assignments and from the advance notice requirement. It does not allow derogation from the additional-hours limit in subsection 2 or from the reference time window. The limit is 24 × 1.25 = 30 hours. The Friday assignment is binding and brings the total to 30 hours. Saturday falls outside the reference time window and is not binding, even as a substitute. For the full week of illness, subsection 4 excludes the three excluded weeks and their corresponding time from the calculation. The result is 294 ÷ 10 = 29.4 hours, or €588.

\paragraph{Scored propositions.}\begin{enumerate}
\item The collective agreement’s two-hour assignment rule is valid. Friday falls within the reference time window and was notified on time, so the call-out is permissible.
\item The 40\% extension is ineffective. No more than 30 hours per week may be required unilaterally.
\item The Saturday call-out is not binding because it is outside the reference days.
\item Sick pay is €588 for 29.4 hours after the three weeks are excluded.
\end{enumerate}

\paragraph{Evidence and conditional branches.} The three E1 units come from \href{https://www.gesetze-im-internet.de/tzbfg/__12.html}{§12 of the German Part-Time and Fixed-Term Employment Act (TzBfG)}. Subsections 1, 3 and 6 establish the scheduling rules and permissible collective derogations. Subsection 2 sets the additional-hours ceiling. Subsection 4 governs averaging for sick pay. E2 interleaves these German units with one sickness-related excerpt from each of Japan, Brazil, France, Saudi Arabia, the Netherlands, mainland China and the United States. E3 removes only subsection 4. The numerical facts permit arithmetic under either averaging rule, so the surviving sources cannot establish which weeks to exclude. P1–P3 retain full support in E3. P4 loses support. Friday and Saturday test the reference window independently of the weekly cap. The sick-pay branch asks for a different legal operation on a later week.

\paragraph{Sol, E2.} “Under § 12(4) TzBfG, continued pay is calculated using average working time over the preceding three months. Periods of annual leave and short-time work are excluded from that calculation.” Sol reaches the 30-hour ceiling and €588 under two different subsections. All four legal propositions receive content and grounding credit.

\paragraph{Sol, E3.} “The supplied extracts do \emph{not} include the specific sick-pay averaging provision.” Sol keeps the same 30-hour ceiling and €588 and identifies its external basis for the sick-pay calculation. All four propositions receive content credit, while only three receive grounding credit.